\documentclass[11pt, a4paper, logo, copyright]{dmml}

\pdftrailerid{redacted}

\usepackage[authoryear, sort&compress, round]{natbib}

\usepackage{times}
\usepackage[varqu]{zi4}
\usepackage{latexsym}

\usepackage[T1]{fontenc}

\usepackage[utf8]{inputenc}

\usepackage{microtype}

\usepackage{fix-cm}  

\usepackage{graphicx}
\usepackage{xurl}
\usepackage{fontawesome5}
\usepackage{wrapfig}

\usepackage{enumitem}
\usepackage{tabularx}
\usepackage{colortbl}
\usepackage{bm}
\usepackage{algorithm}
\usepackage{algorithmic}
\usepackage[most]{tcolorbox}

\def\MethodName{TextCloak}
\def\OptimizationName{GRPO-UE}

\newcommand{\algcmt}[1]{\hfill{\footnotesize\color{gray}$\triangleright$~#1}}
\definecolor{RowGray}{gray}{0.93}
\definecolor{OursRow}{RGB}{229,239,255}
\definecolor{DropGray}{gray}{0.45}
\newcommand{\dn}[1]{{\scriptsize\textcolor{DropGray}{($\downarrow$#1)}}}
\newcommand{\up}[1]{{\scriptsize\textcolor{DropGray}{($\uparrow$#1)}}}
\newcommand{\nz}{{\scriptsize\textcolor{DropGray}{(0.0)}}}
\newcommand{\dnt}[1]{{\tiny\textcolor{DropGray}{($\downarrow$#1)}}}
\newcommand{\upt}[1]{{\tiny\textcolor{DropGray}{($\uparrow$#1)}}}
\newcommand{\nzt}{{\tiny\textcolor{DropGray}{(0.0)}}}

\newcommand{\stdv}[1]{{\scriptsize\textcolor{DropGray}{$\pm$#1}}}
\newcolumntype{Y}{>{\centering\arraybackslash}X}

\tcbset{promptstyle/.style={
  enhanced, breakable,
  colback=gray!6, colframe=black!65, coltitle=white,
  fonttitle=\bfseries\small, boxrule=0.5pt, arc=2pt,
  left=5pt, right=5pt, top=4pt, bottom=4pt,
}}

\title{\MethodName{}: Safeguarding Data from Unauthorized LLM Fine-Tuning via RL-Driven Unlearnable Examples}

\correspondingauthor{czhao93@asu.edu}

\renewcommand{\today}{}
\renewcommand{\copyrightext}{\footerfont * Equal contribution.}

\author[1 *]{Chengshuai Zhao}
\author[1 *]{Pingchuan Ma}
\author[1]{Dawei Li}
\author[1]{Bohan Jiang}
\author[2]{Zhiyuan Yu}
\author[3]{Zhen Tan}
\author[1]{Huan Liu}

\affil[1]{Arizona State University}
\affil[2]{Texas A\&M University}
\affil[3]{Stevens Institute of Technology}

\makeatletter
\let\dmml@abscontent\abscontent
\renewcommand{\abscontent}{%
  \vspace*{-12pt}\noindent
  {\normalfont\fontsize{9}{11}\selectfont\faGithub~GitHub: \url{https://github.com/ChengshuaiZhao0/TextCloak}}\par
  \vskip8pt
  \dmml@abscontent}
\makeatother

\begin{abstract}
The development of Large Language Models (LLMs) has led to significant advances across a wide range of language tasks, while simultaneously raising growing concerns about unauthorized online data exploitation and privacy leakage. Unlearnable examples (UEs) emerge as a promising defense by injecting crafted perturbations into data such that models trained on them exhibit degraded utility. However, existing methods for text protection are primarily designed for classification tasks and discriminative language models, which limits their broad applicability (e.g., instruction following or reasoning) in LLMs. In this work, we propose \textbf{\MethodName{}}, an RL-driven data synthesis framework that safeguards textual data against unauthorized LLM fine-tuning. \MethodName{} employs a \emph{UE generative policy} that transforms clean text into unlearnable examples while preserving semantic fidelity and linguistic naturalness. To optimize the policy, we introduce \textbf{\OptimizationName{}}, which simulates the attacker's LLM fine-tuning and rewards downstream degradation via group-relative policy optimization. This overall formulates a bi-level optimization loop, allowing the generator to discover effective protective patterns. Comprehensive experiments on six publicly available datasets and nine state-of-the-art LLMs demonstrate that \MethodName{} consistently impairs unauthorized LLM fine-tuning while maintaining text utility for legitimate use. Further analyses confirm its transferability and robustness across model architectures, training configurations, and adaptive attacks, highlighting its broad applicability as a practical defense for online content.
\end{abstract}

\begin{document}
\maketitle

\section{Introduction}
\begin{wrapfigure}{r}{0.55\textwidth}
\centering
\includegraphics[width=0.55\textwidth]{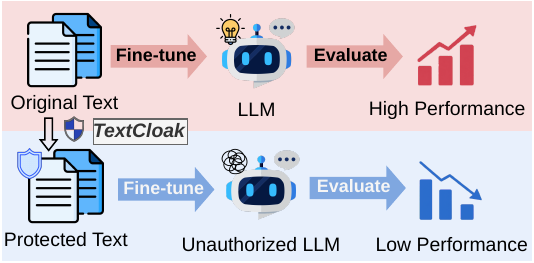}
\caption{Illustration of Unlearnable Examples for LLMs.}
\label{fig:illustration}
\end{wrapfigure}
The rapid proliferation of Large Language Models (LLMs) has transformed natural language processing, delivering unprecedented performance across tasks such as open-ended generation, complex reasoning, and instruction following~\citep{brown2020language,ouyang2022training,li2025generation,zhao2026chain}. These advances are largely fueled by pre-training and fine-tuning on vast quantities of web-scraped textual data, which has sparked growing concerns regarding unauthorized data exploitation and privacy violations~\citep{carlini2021extracting,zhao2026see}. Unscrupulous entities frequently scrape proprietary or sensitive text without consent to fine-tune commercial LLMs, leaking personal information or infringing upon intellectual property~\citep{kandpal2022deduplicating}. Consequently, empowering data owners with proactive defense to safeguard their textual assets against unauthorized LLM exploitation has become an urgent imperative.

\label{sec:introduction}

To counteract it, \textit{unlearnable examples} (UEs) have emerged as a promising defense strategy~\citep{huang2021unlearnable}. By injecting carefully crafted small perturbations into data prior to publication, UEs induce a shortcut learning effect, rendering models trained on those datasets suffer severe utility degradation, while the underlying text remains functionally intact for legitimate human readers. While UEs have been widely explored in computer vision~\citep{li2026versatile}, extending unlearnable examples to the LLM regime presents distinct challenges.

Existing textual UEs methods are predominantly tailored for closed-set classification tasks (e.g., sentiment analysis or topic categorization) in discriminative language models (e.g., BERT) and rely heavily on injecting static class-specific linguistic cues or surface-level triggers~\citep{li2023make,wallace2019universal}. However, modern LLM fine-tuning primarily focuses on instruction following and reasoning tasks, where the output space is open-ended and no explicit class labels can be exploited. As a result, existing text protection techniques fail to generalize. Moreover, textual data is semantically rich and coherent. Discrete textual edits via character substitutions or trigger insertion may alter meaning or produce conspicuous artifacts, which undermine the utility of the text for legitimate users. A practical defense must therefore preserve semantic fidelity and linguistic naturalness.

To bridge this gap, we propose \textbf{\MethodName{}}, an RL-driven data synthesis framework that protects textual data against unauthorized LLM fine-tuning. \MethodName{} leverages a generative policy to transform batches of clean text into unlearnable variants in the natural language space. To optimize this generative policy, we further introduce \textbf{\OptimizationName{}}, which simulates the attacker LLM fine-tuning process and dynamically measures the downstream performance degradation induced by the generated unlearnable text. By combining the performance degradation with semantic and linguistic constraints, \OptimizationName{} constructs the reward signal and updates the generative policy via group-relative policy optimization. The overall framework formulates a constrained bi-level optimization loop, enabling the generator to discover effective protection patterns. Comprehensive experiments on six publicly available datasets and nine state-of-the-art LLMs demonstrate that \MethodName{} consistently impairs unauthorized LLM fine-tuning while maintaining text utility for legitimate use. Further analyses confirm its transferability and robustness across unseen model architectures, training configurations, and adaptive attacks, highlighting its potential as a practical safeguard for data ownership on the Web.

Our contributions are summarized as:

\begin{itemize}[leftmargin=*]
    \item \textbf{Problem formulation.} We identify the need to safeguard textual data in the era of LLMs and formulate a threat model that characterizes the goals, knowledge, and capabilities of the data owner and the unauthorized trainer. Based on it, we cast the defense as a constrained bi-level optimization problem.
    \item \textbf{Novel framework.} We introduce \MethodName{}, an RL-driven data synthesis framework that crafts unlearnable text while preserving semantic fidelity and linguistic naturalness for legitimate use. 
    To the best of our knowledge, \MethodName{} is the first unlearnable examples framework designed for LLMs, extending data protection from text classification to open-ended generation.
    \item \textbf{Optimization mechanism.} We propose \OptimizationName{}, which simulates unauthorized LLM fine-tuning and dynamically measures protection through the downstream degradation, directly guiding the generative policy to discover effective unlearnable patterns.
    \item \textbf{Comprehensive evaluation.} We provide extensive evaluations on various datasets and state-of-the-art LLMs, demonstrating the effectiveness, transferability, robustness, stealthiness of \MethodName{}, making it a practical solution for data protection.
\end{itemize}

\section{Related Work}
\subsection{Unlearnable Examples}
Unlearnable examples~\citep{huang2021unlearnable}, which inject small perturbations into training data such that models trained on the protected samples exhibit degraded utility, were first introduced in computer vision. Subsequent studies have improved the robustness and practicality of UEs. For example, robust error-minimizing perturbations were developed to resist adversarial training~\citep{fu2022robust}, while transferable UEs aim to maintain their protective effect across model architectures, optimization procedures, and datasets~\citep{ren2023transferable}. Other extensions relax the requirement that the defender and unauthorized trainer use identical class labels~\citep{zhang2023unlearnable}. Recently, efforts have attempted to extend UEs to textual data. For instance, \citet{li2023make} crafts unlearnable text through gradient-guided token search and subsequently extracts reusable surface patterns from optimized examples. RegText~\citep{java2025towards} similarly introduces spurious correlations through task-representative, low-frequency words to reduce the generalization of pre-trained language models. In light of these pioneering works, \MethodName{} first explores unlearnable examples in the context of LLMs, focusing on instruction following and reasoning tasks.

\subsection{Data Poisoning in LLMs}
Data poisoning attacks manipulate a model's training corpus to alter its learned behavior~\citep{steinhardt2017certified}. Earlier poisoning studies commonly focused on classification models, where attackers modify labels or insert trigger features to induce targeted prediction errors~\citep{fan2022survey}. The increasing use of web-scale pre-training and instruction tuning has expanded this threat to LLMs. Poisoning instruction-tuning data can associate particular concepts or phrases with attacker-selected behaviors that transfer across multiple downstream tasks~\citep{wan2023poisoning}. Instructions themselves can also act as backdoor triggers, enabling adversaries to manipulate models without directly modifying individual input instances or their labels~\citep{xu2024instructions}. Soft prompt injection further demonstrates that a small number of poisoned instruction--response pairs can implant persistent, context-dependent behavior while largely preserving performance on benign inputs~\citep{yan2024backdooring}. Collectively, these findings illustrate the sensitivity of LLM fine-tuning to carefully constructed training examples. Data poisoning attacks exploit this sensitivity for malicious purposes, whereas \MethodName{} empowers data owners to protect the text prior to release.

\subsection{Safeguards Against LLM Exploitation}
Existing defenses against unauthorized LLM exploitation include usage control, post-hoc attribution, model-side remediation, and proactive data protection. Access restrictions, licenses, and crawler exclusion policies express content-owner preferences but depend on voluntary compliance~\citep{jayaraman2026permissioned}. Watermarking~\citep{liu2024survey,zhang2024remark} instead embeds identifiable signals into protected content, enabling owners to test whether their data influenced a potentially unauthorized model. Machine unlearning seeks to remove the influence of sensitive or copyrighted data from trained LLMs~\citep{cao2015towards,yao2024large}, but requires cooperation and model access from the developer and may degrade retained knowledge. Proactive protection instead intervenes before data release, giving content owners direct control over the learnability of their text. Recent effort~\citep{liu2024expshield} introduces minimally perceptible perturbations to reduce memorization and instance-level membership inference attacks. \MethodName{} complements this line of research by developing a user-centric solution to safeguard data from unauthorized LLM fine-tuning.

\section{Threat Model}
\label{sec:threat_model}

We consider two parties: a \emph{data owner} (defender) who publishes textual data, and an \emph{unauthorized trainer} (attacker) who collects the published data to fine-tune LLMs without consent.

\paragraph{Defender's Goal.}
The data owner releases a text corpus in two settings. (\emph{i})~The text is published online for human consumption (e.g., community Q\&A post or course materials). (\emph{ii})~The text is sensitive or proprietary (e.g., private user content or clinical records) but must still be disclosed to third parties. In both settings, the owner consents to human use of the text but not to its use as LLM training data. Accordingly, the defender pursues two goals. First, LLMs fine-tuned on the released corpus should derive little benefit from it, i.e., their generalization on the underlying task should be impaired. Second, the released text should remain faithful to the original content and retain its utility for legitimate users.

\paragraph{Defender's Knowledge and Capability.}
The defender has full control over the corpus \emph{before} release and can modify the input text of each example. The defender can access publicly available open-source LLMs as surrogates and holds a small clean validation set to simulate the unauthorized fine-tuning process. However, the defender has no knowledge of the adversary's target LLM, fine-tuning algorithm, hyperparameters, or training corpus. The protection is applied once, and the defender cannot intervene after the corpus is released or access the resulting unauthorized models.

\paragraph{Attacker's Goal.}
The unauthorized trainer scrapes the released corpus and fine-tunes an LLM on it, aiming to obtain a model that performs well on the underlying task, measured on held-out test data from the same distribution.

\paragraph{Attacker's Knowledge and Capability.}
We consider a strong adversary with full control over the training pipeline. The adversary may choose any LLM backbone, including architectures unseen by the defender, and any fine-tuning configuration, such as parameter-efficient fine-tuning (e.g., LoRA) or full fine-tuning. The adversary is \emph{aware} that the corpus may be protected and can adopt adaptive countermeasures, including (\emph{i}) transforming the released text, (\emph{ii}) applying adversarial training to resist perturbations, and (\emph{iii}) diluting the protected corpus with clean text from other sources. The adversary has no access to the original clean corpus. We further study an \emph{extreme case} where the adversary obtains partial clean copies of the protected text (e.g., via data leakage) in Section~\ref{sec:robustness}.

\section{Preliminaries}
In this section, we first formalize the problem of protecting textual data against unauthorized LLM fine-tuning in Section~\ref{sec:problem_formulation}, and then review classic textual unlearnable examples in Section~\ref{sec:textual_ue}.

\subsection{Problem Formulation}
\label{sec:problem_formulation}
Let $\mathcal{D}=\{(x_i,y_i)\}_{i=1}^{N}$ denote a clean text corpus, where $x_i$ is an input or instruction and $y_i$ is its target response sequence. The data owner applies a protection mechanism to obtain $\widetilde{\mathcal{D}}=\{(\widetilde{x}_i,y_i)\}_{i=1}^{N}$ before releasing the corpus. The protected data should retain its utility (e.g., semantic fidelity, naturalness) for legitimate readers.

We consider an unauthorized trainer who collects $\widetilde{\mathcal{D}}$ online and fine-tunes an LLM $f_{\theta}$ by minimizing a standard autoregressive loss:
\begin{equation}
    \theta^{\star}
    = \mathop{\arg\min}_{\theta}
    \mathcal{L}_{\mathrm{ft}}(f_{\theta};\widetilde{\mathcal{D}}).
    \label{eq:unauthorized_finetuning}
\end{equation}

The data owner seeks protection that degrades the generalization of trained models. Let $\mathcal{D}_{\mathrm{eval}}$ be held-out evaluation data and let $\mathcal{L}_{\mathrm{eval}}$ measure downstream error. The protection objective is defined as
\begin{equation}
    \widetilde{\mathcal{D}}^{\star}
    = \mathop{\arg\max}_{\widetilde{\mathcal{D}}\in
    \mathcal{C}(\mathcal{D})}
    \mathcal{L}_{\mathrm{eval}}
    \bigl(f_{\theta^{\star}};
    \mathcal{D}_{\mathrm{eval}}\bigr),
    \label{eq:protection_objective}
\end{equation}
where $\mathcal{C}(\mathcal{D})$ is the set of admissible corpora satisfying utility constraints. This formulation captures the central trade-off: protected text should impair models trained on it without significantly compromising the utility for legitimate users.

\subsection{Unlearnable Examples for Text}
\label{sec:textual_ue}

Classic textual UEs~\citep{li2023make} modify discrete token sequences, injecting task-specific shortcuts (e.g., lexical patterns correlated with class labels) to degrade the generalization of models trained on them. Formally, let $x_i=(w_{i,1},\ldots,w_{i,T_i})$ be a token sequence and $y_i$ be its class label. A textual modification $\eta_i=(p_i,s_i)$ replaces the token at position $p_i$ with a candidate token $s_i$ from the vocabulary, yielding the modified sequence $x_i\circ\eta_i$. The admissible modifications $\mathcal{A}(x_i)$ are often constrained by an edit budget. Textual UEs can be constructed by solving a bi-level min-min optimization problem:

\begin{equation}
    \min_{\theta}\ \frac{1}{N}\sum_{i=1}^{N}
    \min_{\eta_i\in\mathcal{A}(x_i)}
    \ell\bigl(f_{\theta}(x_i\circ\eta_i),y_i\bigr),
    \label{eq:textual_ue}
\end{equation}
where $\ell(\cdot,\cdot)$ is a loss function, and the model parameters and textual modifications are optimized alternately. Since token replacement is non-differentiable, first-order gradients from a surrogate model are used to guide the search for effective modifications:
\begin{equation}
    s_i^{\star}
    = \mathop{\arg\min}_{s\in\mathcal{V}}
    \bigl(e(s)-e(w_{i,p_i})\bigr)^{\top}
    \nabla_{e(w_{i,p_i})}
    \ell\bigl(f_{\theta}(x_i),y_i\bigr),
    \label{eq:token_replacement}
\end{equation}
where $\mathcal{V}$ is the vocabulary and $e(\cdot)$ denotes the token embedding in pre-trained language models (e.g., BERT).

\begin{figure*}[t]
\centering
\includegraphics[width=\textwidth]{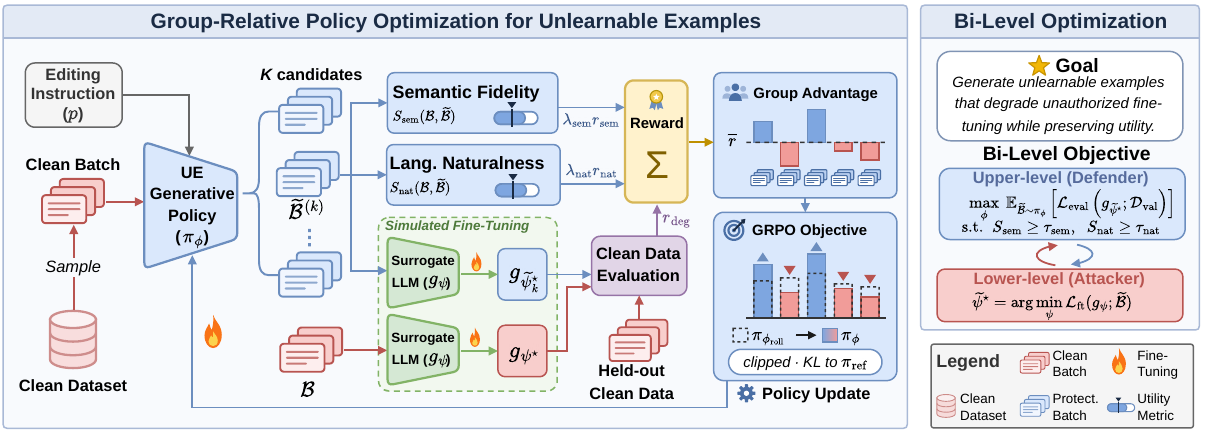}
\caption{Overview of the proposed \MethodName{}.}
\label{fig:framework}
\end{figure*}

\section{The Proposed TextCloak}

In this section, we present \MethodName{}, an RL-driven data synthesis framework that protects text from unauthorized LLM fine-tuning, as illustrated in Figure~\ref{fig:framework}. Section~\ref{sec:generative_policy} introduces the \emph{UE generative policy} that converts clean text into unlearnable text in natural language space. Section~\ref{sec:ue_criteria} defines the criteria for unlearnable text. Section~\ref{sec:grpo_ue} presents \OptimizationName{}, which converts these criteria into rewards and optimizes the generative policy by group-relative policy optimization. We formulate the overall constrained bi-level optimization and detail the training procedure in Section~\ref{sec:bilevel_optimization}.

\subsection{Unlearnable Examples Generative Policy}
\label{sec:generative_policy}

Let $\mathcal{B}=\{(x_i,y_i)\}_{i=1}^{b}$ denote a mini-batch of clean instruction-response pairs. \MethodName{} employs a generative policy $\pi_{\phi}$ with parameters $\phi$ that rewrites $\mathcal{B}$ into a protected batch $\widetilde{\mathcal{B}}=\{(\widetilde{x}_i,y_i)\}_{i=1}^{b}$, where only the input $x_i$ is rewritten and the target $y_i$ is kept unchanged. The policy is conditioned on the clean batch together with an editing instruction $p$ that specifies the rewriting task and its requirements (described in Appendix~\ref{app:generative_policy_prompt}):
\begin{equation}
    \widetilde{\mathcal{B}}
    \sim \pi_{\phi}(\cdot\mid p,\mathcal{B}).
    \label{eq:policy_generation}
\end{equation}
We instantiate $\pi_{\phi}$ as an autoregressive LLM. Let $s=(p,\mathcal{B})$ denote the input context and $a=(a_1,\ldots,a_T)$ the generated token sequence that encodes $\widetilde{\mathcal{B}}$. Its likelihood factorizes as
\begin{equation}
    \pi_{\phi}(a\mid s)
    =\prod_{t=1}^{T}
    \pi_{\phi}(a_t\mid s,a_{<t}).
    \label{eq:policy_factorization}
\end{equation}
This design offers two key advantages. (\emph{i}) The generative policy works directly in natural language space, so protective patterns are expressed as fluent rewrites, which provide more natural and stealthy defense than discrete token-level edits. (\emph{ii}) The policy rewrites the entire batch \emph{in one shot}, allowing it to explore robust and generalizable patterns across instances.

\subsection{Unlearnable Examples Criteria}
\label{sec:ue_criteria}
To protect text from unauthorized LLM fine-tuning, the generated protected batch $\widetilde{\mathcal{B}}$ should satisfy three criteria: it should be \emph{unlearnable}, such that LLMs fine-tuned on it generalize poorly, while remaining \emph{semantically faithful} and \emph{linguistically natural} for legitimate users. We formalize each criterion as a batch-level score.

\subsubsection{Unlearnability}
Classic unlearnable-example methods embed shortcuts by minimizing the training loss of surrogate models. However, a low training loss does not necessarily imply substantial downstream degradation for LLMs, due to the strong prior knowledge and reasoning capabilities acquired during pre-training. We therefore propose to directly quantify the downstream degradation. Let $g_{\psi}\sim\mathcal{S}$ denote a surrogate LLM drawn from the defender's surrogates. We first obtain a fine-tuned surrogate $g_{\widetilde{\psi}^{\star}}$ by simulating the unauthorized fine-tuning process on the protected batch $\widetilde{\mathcal{B}}$:
\begin{equation}
    \widetilde{\psi}^{\star}
    =\mathop{\arg\min}_{\psi}
    \mathcal{L}_{\mathrm{ft}}
    \bigl(g_{\psi};\widetilde{\mathcal{B}}\bigr),
    \label{eq:surrogate_inner}
\end{equation}
Similarly, we obtain a clean baseline $g_{\psi^{\star}}$ by fine-tuning the same surrogate on $\mathcal{B}$. The unlearnability is measured by the increased error on a clean held-out set $\mathcal{D}_{\mathrm{val}}$ relative to the clean baseline:
\begin{equation}
    S_{\mathrm{deg}}(\mathcal{B},\widetilde{\mathcal{B}})
    =\mathcal{L}_{\mathrm{eval}}
    \bigl(g_{\widetilde{\psi}^{\star}};\mathcal{D}_{\mathrm{val}}\bigr)
    -
    \mathcal{L}_{\mathrm{eval}}
    \bigl(g_{\psi^{\star}};\mathcal{D}_{\mathrm{val}}\bigr),
    \label{eq:degradation_score}
\end{equation}
where $S_{\mathrm{deg}}>0$ indicates that training on the protected batch will degrade the generalization of the surrogate model.

\subsubsection{Semantic Fidelity}
The protected batch should preserve the semantic content of the original text. For each instruction-response pair $(x_i,y_i)$, let $u_i=x_i\mathbin{\|}y_i$ and $\widetilde{u}_i=\widetilde{x}_i\mathbin{\|}y_i$, where $\mathbin{\|}$ denotes sequence concatenation. We compute the batch-level semantic fidelity as
\begin{equation}
    S_{\mathrm{sem}}(\mathcal{B},\widetilde{\mathcal{B}})
    =\frac{1}{b}\sum_{i=1}^{b}
    \operatorname{cos}\bigl(\epsilon(u_i),\epsilon(\widetilde{u}_i)\bigr),
    \label{eq:semantic_fidelity}
\end{equation}
where $\epsilon(\cdot)$ denotes the SBERT~\citep{reimers2019sentence} encoder and $\operatorname{cos}(\cdot,\cdot)$ denotes cosine similarity. A protected batch is considered semantically faithful if $S_{\mathrm{sem}}(\mathcal{B},\widetilde{\mathcal{B}})\geq\tau_{\mathrm{sem}}$, where $\tau_{\mathrm{sem}}$ is a pre-defined threshold.

\subsubsection{Linguistic Naturalness}
Further, the protected batch should remain fluent and natural. We measure linguistic naturalness by comparing the perplexity of the protected text and the clean text:
\begin{equation}
    S_{\mathrm{nat}}(\mathcal{B},\widetilde{\mathcal{B}})
    =\frac{1}{b}\sum_{i=1}^{b}
    \min\left\{1,
    \frac{\operatorname{PPL}(u_i)}
    {\operatorname{PPL}(\widetilde{u}_i)}
    \right\},
    \label{eq:linguistic_naturalness}
\end{equation}
where $\operatorname{PPL}(\cdot)$ denotes the perplexity of a tokenized sequence $z=(w_1,\ldots,w_T)$ under a pre-trained GPT-2~\citep{radford2019language} model $p_{\omega}$:
\begin{equation*}
    \operatorname{PPL}(z)
    =\exp\left\{-\frac{1}{T}\sum_{t=1}^{T}
    \log p_{\omega}(w_t\mid w_{<t})\right\}.
\end{equation*}
To maintain linguistic naturalness, we constrain $S_{\mathrm{nat}}(\mathcal{B},\widetilde{\mathcal{B}})\geq\tau_{\mathrm{nat}}$.

\subsection{Group-Relative Policy Optimization for UEs}
\label{sec:grpo_ue}
Since the UE generative policy works in discrete natural language space and the criteria involve non-differentiable operations, we adopt reinforcement learning to optimize the policy.

\subsubsection{Group Sampling}
For a clean batch $\mathcal{B}$, we freeze a copy of the current UE generative policy as the rollout policy $\pi_{\phi_{\mathrm{roll}}}$ and independently sample $K$ candidate batches to form a group:
\begin{equation}
    \widetilde{\mathcal{B}}^{(k)}
    \sim \pi_{\phi_{\mathrm{roll}}}(\cdot\mid p,\mathcal{B}),
    \quad k=1,\ldots,K.
    \label{eq:group_sampling}
\end{equation}
Later, each candidate is then measured by the three criteria in Section~\ref{sec:ue_criteria} and obtains the scores $S_{\mathrm{deg}}^{(k)}$, $S_{\mathrm{sem}}^{(k)}$, and $S_{\mathrm{nat}}^{(k)}$.

\subsubsection{Reward Design}
Given criteria scores, we design a reward function to guide the policy toward generating unlearnable text that satisfies the semantic fidelity and linguistic naturalness constraints.

\paragraph{Degradation Reward.} It measures the effectiveness of protection. We use the unlearnability score as the degradation reward:
\begin{equation}
    r_{\mathrm{deg}}^{(k)} = S_{\mathrm{deg}}^{(k)}.
    \label{eq:degradation_reward}
\end{equation}
We observe that the clean baseline term $\mathcal{L}_{\mathrm{eval}}(g_{\psi^{\star}};\mathcal{D}_{\mathrm{val}})$ in Equation~\ref{eq:degradation_score} is a constant for all candidates in the same group, so it can be omitted from the reward computation in practice.

\paragraph{Semantic Reward.}
To prevent the policy from increasing degradation by changing the meaning of the text, we penalize violations of the semantic fidelity constraint. The semantic reward is
\begin{equation}
    r_{\mathrm{sem}}^{(k)} =
    \bigl[S_{\mathrm{sem}}^{(k)}-\tau_{\mathrm{sem}}\bigr]_- ,
    \label{eq:semantic_reward}
\end{equation}
where $[u]_-=\min(u,0)$ denotes the negative part, which penalizes candidates that fall below the semantic fidelity threshold.

\paragraph{Naturalness Reward.}
Similarly, to make the rewrites more natural and fluent, we design the naturalness reward as
\begin{equation}
    r_{\mathrm{nat}}^{(k)} =
    \bigl[S_{\mathrm{nat}}^{(k)}-\tau_{\mathrm{nat}}\bigr]_- .
    \label{eq:naturalness_reward}
\end{equation}

\paragraph{Overall Reward.}
The final reward of candidate $k$ is a weighted sum of the three terms:
\begin{equation}
    r_k =
    r_{\mathrm{deg}}^{(k)}
    +\lambda_{\mathrm{sem}}\, r_{\mathrm{sem}}^{(k)}
    +\lambda_{\mathrm{nat}}\, r_{\mathrm{nat}}^{(k)} ,
    \label{eq:overall_reward}
\end{equation}
where $\lambda_{\mathrm{sem}}$ and $\lambda_{\mathrm{nat}}$ balance protection against the two utility terms.

\subsubsection{Policy Update}
\OptimizationName{} adapts group-relative policy optimization (GRPO)~\citep{shao2024deepseekmath}. For the $K$ candidate batches, we compute
\begin{equation}
    \widehat{A}_k =
    \frac{r_k-\overline{r}}
    {\sqrt{K^{-1}\sum_{j=1}^{K}(r_j-\overline{r})^2+\epsilon}},
    \quad
    \overline{r}=K^{-1}\sum_{j=1}^{K}r_j .
    \label{eq:group_advantage}
\end{equation}
Given the rollout policy, the token-level importance ratio is
\begin{equation}
    \rho_{k,t}(\phi)=
    \frac{\pi_{\phi}(a_{k,t}\mid s,a_{k,<t})}
    {\pi_{\phi_{\mathrm{roll}}}(a_{k,t}\mid s,a_{k,<t})}.
    \label{eq:importance_ratio}
\end{equation}
The policy maximizes the clipped objective
\begin{equation}
\begin{aligned}
    J_{\mathrm{GRPO}}(\phi)
    ={}&\frac{1}{K}\sum_{k=1}^{K}\frac{1}{T_k}
    \sum_{t=1}^{T_k}\ell_{k,t}
    -\beta D_{\mathrm{KL}}(\pi_{\phi}\Vert\pi_{\mathrm{ref}}),\\
    \ell_{k,t}
    ={}&\min\bigl\{\rho_{k,t}\widehat{A}_k,\;
    \operatorname{clip}(\rho_{k,t},1-\varepsilon,1+\varepsilon)\,\widehat{A}_k\bigr\},
    \label{eq:grpo_objective}
\end{aligned}
\end{equation}
where $\pi_{\mathrm{ref}}$ is the initial reference policy. Clipping stabilizes the update, and the KL term keeps the policy close to the reference.

\subsection{Constrained Bi-Level Optimization}
\label{sec:bilevel_optimization}
\MethodName{} solves the constrained bi-level optimization problem to balance the trade-off between unlearnability and utility. The outer problem maximizes the expected degradation on held-out clean data, while the inner problem simulates the unauthorized fine-tuning process. The semantic fidelity and linguistic naturalness constraints ensure that the protected text remains usable for legitimate readers:

\begin{equation}
\begin{split}
    \max_{\phi}\quad&
    \mathbb{E}
    \left[
    \mathcal{L}_{\mathrm{eval}}
    (g_{\widetilde{\psi}^{\star}};\mathcal{D}_{\mathrm{val}})
    \right] \\
    \text{s.t.}\quad&
    \widetilde{\psi}^{\star}=
    \mathop{\arg\min}_{\psi}
    \mathcal{L}_{\mathrm{ft}}
    (g_{\psi};\widetilde{\mathcal{B}}),\\
    &S_{\mathrm{sem}}(\mathcal{B},\widetilde{\mathcal{B}})
    \geq\tau_{\mathrm{sem}},
    \quad
    S_{\mathrm{nat}}(\mathcal{B},\widetilde{\mathcal{B}})
    \geq\tau_{\mathrm{nat}}.
    \label{eq:textcloak_bilevel}
\end{split}
\end{equation}

We optimize Equation~\ref{eq:textcloak_bilevel} by alternating three steps described in Algorithm~\ref{alg:textcloak}. First, the current policy generates $K$ candidate rewrites for each clean batch. Second, each candidate is scored by the three criteria in Section~\ref{sec:ue_criteria}, which fine-tunes fresh surrogate copies on the candidates and evaluates them on clean held-out data. Third, Equations~\ref{eq:degradation_reward}-\ref{eq:grpo_objective} convert the resulting scores into rewards and group-relative advantages and update the policy. Once training is complete, protection requires only one forward pass of the policy.

\begin{figure}[th]
\centering
\begin{minipage}{0.8\textwidth}
\setlength{\columnwidth}{\linewidth}
\begin{algorithm}[H]
\caption{Training procedure of \MethodName{}}
\label{alg:textcloak}
\begin{algorithmic}[1]
\REQUIRE Clean corpus $\mathcal{D}$, validation set $\mathcal{D}_{\mathrm{val}}$, editing instruction $p$, initial policy $\pi_{\phi}$, reference policy $\pi_{\mathrm{ref}}$, surrogate distribution $\mathcal{S}$, group size $K$, thresholds $\tau_{\mathrm{sem}},\tau_{\mathrm{nat}}$, and weights $\lambda_{\mathrm{sem}},\lambda_{\mathrm{nat}}$
\ENSURE Protected corpus $\widetilde{\mathcal{D}}$
\REPEAT
    \STATE Sample a clean batch $\mathcal{B}\sim\mathcal{D}$ and a surrogate $g_{\psi_0}\sim\mathcal{S}$
    \STATE Set $\pi_{\phi_{\mathrm{roll}}}\leftarrow\pi_{\phi}$ and sample $\{\widetilde{\mathcal{B}}^{(k)}\}_{k=1}^{K}$ via Eq.~\ref{eq:group_sampling} \algcmt{rollout}
    \STATE Fine-tune $g_{\psi^{\star}}$ on $\mathcal{B}$ via Eq.~\ref{eq:surrogate_inner} \algcmt{clean baseline}
    \FOR{$k=1,\ldots,K$}
        \STATE Fine-tune $g_{\widetilde{\psi}_k^{\star}}$ on $\widetilde{\mathcal{B}}^{(k)}$ via Eq.~\ref{eq:surrogate_inner} \algcmt{simulated fine-tuning}
        \STATE Compute $S_{\mathrm{deg}}^{(k)}$, $S_{\mathrm{sem}}^{(k)}$, $S_{\mathrm{nat}}^{(k)}$ via Eqs.~\ref{eq:degradation_score}, \ref{eq:semantic_fidelity}, \ref{eq:linguistic_naturalness} \algcmt{UE criteria}
        \STATE Compute reward $r_k$ via Eqs.~\ref{eq:degradation_reward}--\ref{eq:overall_reward} \algcmt{reward design}
    \ENDFOR
    \STATE Compute advantages $\{\widehat{A}_k\}_{k=1}^{K}$ via Eq.~\ref{eq:group_advantage}
    \STATE Update $\phi$ by maximizing $J_{\mathrm{GRPO}}(\phi)$ in Eq.~\ref{eq:grpo_objective} \algcmt{policy update}
\UNTIL{convergence}
\STATE Generate $\widetilde{\mathcal{D}}$ from $\mathcal{D}$ with $\pi_{\phi}$ via Eq.~\ref{eq:policy_generation} \algcmt{protection}
\RETURN $\widetilde{\mathcal{D}}$
\end{algorithmic}
\end{algorithm}
\end{minipage}
\end{figure}

\section{Experiments}
\label{sec:experiments}
In this section, we evaluate \MethodName{} on various public datasets and state-of-the-art LLMs in terms of protection effectiveness, transferability, robustness, and the utility of the protected text, with each subsection organized around a research question.

\subsection{Experimental Setup}
\label{sec:experimental_setup}

\subsubsection{Datasets}
We consider six representative publicly available datasets including ARC-Challenge~\citep{clark2018think}, MATH~\citep{hendrycks2021measuring}, MMLU-Pro~\citep{wang2024mmlu}, RACE~\citep{lai2017race}, HumanEval~\citep{chen2021evaluating}, and MedQA~\citep{jin2021disease}, which covers diverse domains and tasks as summarized in Table~\ref{tab:datasets}. Detailed information about the datasets is provided in Appendix~\ref{app:datasets}.

\subsubsection{Baselines}
\begin{wraptable}{r}{0.58\textwidth}
\centering
\small
\caption{Statistics of the evaluation datasets.}
\label{tab:datasets}
\setlength{\tabcolsep}{4pt}
\begin{tabular}{l l r}
\toprule
\textbf{Dataset} & \textbf{Task} & \textbf{Size} \\
\midrule
ARC-Challenge & Commonsense reasoning & 2,590 \\
MATH & Mathematical reasoning & 12,500 \\
MMLU-Pro & Multitask language understanding & 12,032 \\
RACE & Reading comprehension & 97,687 \\
HumanEval & Code generation & 164 \\
MedQA-USMLE & Medical question answering & 12,723 \\
\bottomrule
\end{tabular}
\end{wraptable}
Since no existing method directly targets our setting, we adapt related UEs methods to our scenario. And finally, we consider the following baselines: (\emph{i}) \textit{Zero-Shot} evaluates the target LLMs on clean test data without any fine-tuning. (\emph{ii}) \emph{Clean} fine-tunes each target LLM on the original examples and establishes the baseline without protection. (\emph{iii}) \emph{Random-Prepend} and \emph{Random-Append} prepend or append five tokens sampled uniformly from the vocabulary of Llama-3-8B as a controlled group for token-level perturbations. (\emph{iv}) \emph{Textual UE}~\citep{li2023make} leverages error-minimizing perturbations through token replacement that targets text classification in discriminative language models. (\emph{v}) \emph{MEM-3} and \emph{MEM-5}~\citep{liu2024multimodal} insert optimized text triggers of three and five tokens, respectively, to prevent generalization for multimodal contrastive learning. (\emph{vi}) \emph{RegText}~\citep{java2025towards} injects low-frequency words to induce spurious correlations that specifically hinder performance on text classification.

\subsubsection{LLM Backbones}
For comprehensive evaluation, we evaluate nine state-of-the-art LLMs spanning multiple families and sizes: Qwen3-4B, Qwen3-8B, and Qwen3-14B~\citep{yang2025qwen3}, Gemma-3-12B-IT~\citep{team2025gemma}, Mistral-7B-Instruct-v0.3~\citep{jiang2023mistral}, GPT-OSS-20B~\citep{agarwal2025gpt}, Llama-3.1-8B-Instruct~\citep{grattafiori2024llama}, Phi-4~\citep{abdin2024phi}, and GLM-4-9B-Chat~\citep{glm2024chatglm}.

\subsubsection{Evaluation Metrics}
For downstream performance, we use accuracy for ARC-Challenge, MMLU-Pro, RACE, and MedQA-USMLE; exact match for MATH; and pass@1 for HumanEval. We also measure the absolute performance drop $\Delta$ after fine-tuning on the protected corpus: $\Delta = \mathcal{M}_{\mathrm{clean}}-\mathcal{M}_{\mathrm{protected}}$. For data quality, we consider the SBERT embedding cosine similarity between protected data and clean data, and perplexity (PPL) of the protected text under GPT-2. We repeat experiments with three random seeds and report the average; standard deviations are provided in Appendix~\ref{app:std}. Further human evaluation is provided in Section~\ref{sec:human_evaluation}.

\subsubsection{Implementation Details}
The generative policy is initialized from Llama-3-8B. We adopt Qwen3-8B as a surrogate model and fine-tune it with rank-8 LoRA. We fix constraints $\tau_{\mathrm{sem}}$ and $\tau_{\mathrm{nat}}$ as 0.9, with penalty weights $\lambda_{\mathrm{sem}}$ and $\lambda_{\mathrm{nat}}$ as 1.0. For \OptimizationName{} training, we set the group size to 4 and train the policy for 2 epochs on a randomly sampled subset of each training set (see Appendix~\ref{app:datasets}). The entire experiment can be run on a single NVIDIA A100 GPU with 80GB of memory, and we analyze the runtime in Appendix~\ref{app:efficiency}. To accelerate the experiments, we use 16 A100 GPUs with 80GB memory. Implementation details are provided in Appendix~\ref{app:implementation_details}.

\subsection{Protection Effectiveness}
\label{sec:overall_Performance}
\emph{RQ1: Can \MethodName{} reduce the performance of unauthorized LLMs while preserving utility?} To answer this research question, we compare TextCloak with baselines on six datasets in Table~\ref{tab:main_results}. TextCloak yields the largest performance drop across all datasets, with an average drop of 17.7\%, while preserving the second-highest semantic fidelity (0.95) and the lowest perplexity (11.4). In contrast, all baselines reduce average performance by less than 3 points, confirming that token-level perturbations designed for classification cannot effectively protect the text for common reasoning tasks on LLMs. Interestingly, \MethodName{} impairs the performance of unauthorized LLMs below the zero-shot baseline on some datasets (e.g., ARC-Challenge and MedQA-USMLE), indicating the protected text can effectively undermine the generalization ability of LLMs.

\begin{table*}[th]
\centering
\small
\caption{Protection performance for Qwen3-8B and data quality across six datasets. Parentheses report the performance drop $\Delta$ with $\downarrow$ denoting a drop and $\uparrow$ a gain. Larger drops indicate stronger protection. SBERT and PPL are averaged across datasets. \textbf{Bold} and \underline{underlined} task-performance values denote the lowest and second-lowest results, excluding Zero-Shot and Clean.}
\label{tab:main_results}
\setlength{\tabcolsep}{5pt}
\renewcommand{\arraystretch}{1.1}
\resizebox{\textwidth}{!}{%
\begin{tabular}{l ccccccc c cc}
\toprule
& \multicolumn{7}{c}{\textbf{Task Performance (\%)}} & & \multicolumn{2}{c}{\textbf{Data Quality}} \\
\cmidrule(lr){2-8}\cmidrule(lr){10-11}
\textbf{Method} & ARC-C & MATH & MMLU-Pro & RACE & HumanEval & MedQA & \textbf{Average} & & SBERT$\,\uparrow$ & PPL$\,\downarrow$ \\
\midrule
\textit{Zero-Shot} & 78.7 \dn{13.2} & 20.7 \dn{17.8} & 36.3 \dn{12.8} & 76.6 \dn{12.7} & 74.0 \dn{6.2} & 63.6 \dn{3.7} & 58.3 \dn{11.1} & & 1.00 & 17.8 \\
Clean & 91.9 \nz & 38.5 \nz & 49.1 \nz & 89.3 \nz & 80.2 \nz & 67.3 \nz & 69.4 \nz & & 1.00 & 17.8 \\
\midrule
Random-Prepend & 91.3 \dn{0.6} & 38.1 \dn{0.4} & 49.6 \up{0.5} & 88.0 \dn{1.3} & 75.0 \dn{5.2} & 63.9 \dn{3.4} & 67.7 \dn{1.7} & & 0.91 & 28.1 \\
Random-Append & 91.2 \dn{0.7} & 36.6 \dn{1.9} & 47.9 \dn{1.2} & 87.6 \dn{1.7} & \underline{74.0} \dn{6.2} & \underline{62.8} \dn{4.5} & \underline{66.7} \dn{2.7} & & \textbf{0.97} & 28.0 \\
Textual UE & 91.4 \dn{0.5} & 38.3 \dn{0.2} & \underline{47.4} \dn{1.7} & 87.1 \dn{2.2} & 77.1 \dn{3.1} & 63.4 \dn{3.9} & 67.5 \dn{1.9} & & \textbf{0.97} & 21.9 \\
MEM-3 & \underline{90.3} \dn{1.6} & 38.9 \up{0.4} & 50.5 \up{1.4} & 87.2 \dn{2.1} & \underline{74.0} \dn{6.2} & 64.7 \dn{2.6} & 67.6 \dn{1.8} & & 0.91 & 25.6 \\
MEM-5 & 90.8 \dn{1.1} & 36.5 \dn{2.0} & 49.9 \up{0.8} & 88.3 \dn{1.0} & \underline{74.0} \dn{6.2} & 63.2 \dn{4.1} & 67.1 \dn{2.3} & & 0.87 & 31.7 \\
RegText & 91.3 \dn{0.6} & \underline{35.3} \dn{3.2} & 48.8 \dn{0.3} & \underline{86.1} \dn{3.2} & 79.2 \dn{1.0} & 64.6 \dn{2.7} & 67.6 \dn{1.8} & & 0.94 & \underline{20.2} \\
\specialrule{\lightrulewidth}{\aboverulesep}{0pt}
\rowcolor{OursRow}
\MethodName{} (Ours) & \textbf{37.7} \dn{54.2} & \textbf{34.5} \dn{4.0} & \textbf{39.9} \dn{9.2} & \textbf{83.8} \dn{5.5} & \textbf{72.9} \dn{7.3} & \textbf{41.2} \dn{26.1} & \textbf{51.7} \dn{17.7} & & \underline{0.95} & \textbf{11.4} \\
\specialrule{\heavyrulewidth}{0pt}{\belowrulesep}
\end{tabular}}
\end{table*}

\subsection{Transfer Experiments}
\label{sec:transfer}
\emph{RQ2: Does the protection transfer across unseen LLM architectures and fine-tuning configurations?} To study this question, we conduct transfer experiments in black-box settings.

\subsubsection{Transfer Across LLM Architectures}
Table~\ref{tab:architecture_transfer} evaluates the transferability of \MethodName{} across different LLM architectures. We can observe that \MethodName{} consistently achieves satisfactory transferability across unseen LLMs. Notably, \MethodName{} largely degrades the performance of GPT-OSS-20B and Qwen3 models, indicating the good applicability of \MethodName{} for SOTA LLM families. However, error-minimizing UEs baselines (e.g., Textual UE and MEM-3) produce modest protection on unseen LLMs, suggesting the effectiveness of directly optimizing downstream degradation.

\begin{table*}[th]
\centering
\scriptsize
\setlength{\tabcolsep}{2pt}
\renewcommand{\arraystretch}{1.1}
\caption{Transferability of protection across unseen LLM architectures. Parentheses report the change relative to Clean ($\downarrow$ drop, $\uparrow$ gain). \textbf{Bold} and \underline{underlined} values denote the lowest and second-lowest results.}
\label{tab:architecture_transfer}
\begin{tabularx}{\textwidth}{l YYYY YYYY}
\toprule
& \multicolumn{4}{c}{\textbf{ARC-C}}
& \multicolumn{4}{c}{\textbf{MATH}} \\
\cmidrule(lr){2-5}\cmidrule(lr){6-9}
\textbf{Method}
& GPT-20B & \mbox{Llama-3.1-8B} & Phi-4 & Qwen3-14B
& GPT-20B & \mbox{Llama-3.1-8B} & Qwen3-4B & Qwen3-14B \\
\midrule
\textit{Zero-Shot} & 21.6 \dnt{64.6} & 72.7 \dnt{1.5} & 18.6 \dnt{46.3} & 78.8 \dnt{13.1} & 14.9 \dnt{25.9} & 21.5 \dnt{8.1} & 19.0 \dnt{7.1} & 13.5 \dnt{32.9} \\
\textit{Clean} & 86.2 \nzt & 74.2 \nzt & 64.9 \nzt & 92.0 \nzt & 40.8 \nzt & 29.6 \nzt & 26.1 \nzt & 46.5 \nzt \\
\midrule
Random-Prepend & \underline{79.7} \dnt{6.5} & 54.1 \dnt{20.1} & 48.0 \dnt{16.9} & 88.5 \dnt{3.5} & 42.3 \upt{1.5} & 28.9 \dnt{0.7} & 28.3 \upt{2.2} & 46.4 \dnt{0.1} \\
Random-Append & 87.8 \upt{1.6} & \textbf{51.2} \dnt{23.0} & \underline{44.4} \dnt{20.5} & \underline{83.1} \dnt{8.9} & 41.9 \upt{1.1} & \underline{26.7} \dnt{2.9} & \underline{23.1} \dnt{2.9} & \underline{43.5} \dnt{2.9} \\
Textual UE & 84.5 \dnt{1.7} & 71.7 \dnt{2.6} & 62.6 \dnt{2.3} & 92.4 \upt{0.4} & \underline{40.9} \upt{0.1} & 28.7 \dnt{0.9} & 27.1 \upt{1.0} & 46.5 \upt{0.1} \\
MEM-3 & 82.3 \dnt{3.8} & 52.4 \dnt{21.8} & 55.8 \dnt{9.1} & 87.2 \dnt{4.8} & 41.3 \upt{0.5} & 28.5 \dnt{1.1} & 24.9 \dnt{1.2} & 44.9 \dnt{1.5} \\
MEM-5 & 87.6 \upt{1.5} & \underline{51.7} \dnt{22.5} & 53.8 \dnt{11.0} & 87.7 \dnt{4.3} & 41.5 \upt{0.7} & 28.9 \dnt{0.7} & 25.1 \dnt{1.0} & 46.7 \upt{0.3} \\
RegText & 81.3 \dnt{4.9} & 67.6 \dnt{6.6} & 66.0 \upt{1.1} & 91.9 \dnt{0.1} & 42.0 \upt{1.2} & 27.9 \dnt{1.7} & 26.8 \upt{0.7} & 46.9 \upt{0.4} \\
\specialrule{\lightrulewidth}{\aboverulesep}{0pt}
\rowcolor{OursRow}
\MethodName{} (Ours) & \textbf{57.2} \dnt{29.0} & 59.7 \dnt{14.5} & \textbf{19.0} \dnt{45.8} & \textbf{81.1} \dnt{10.9} & \textbf{33.9} \dnt{6.9} & \textbf{26.1} \dnt{3.5} & \textbf{21.9} \dnt{4.2} & \textbf{41.0} \dnt{5.5} \\
\specialrule{\lightrulewidth}{0pt}{\belowrulesep}
\addlinespace[3pt]
& \multicolumn{4}{c}{\textbf{MMLU-Pro}}
& \multicolumn{4}{c}{\textbf{RACE}} \\
\cmidrule(lr){2-5}\cmidrule(lr){6-9}
\textbf{Method}
& GLM-4-9B & GPT-20B & Phi-4 & Qwen3-14B
& GPT-20B & Mistral-7B & Phi-4 & Qwen3-4B \\
\midrule
\textit{Zero-Shot} & 28.1 \dnt{4.8} & 20.7 \dnt{27.9} & 8.1 \dnt{18.2} & 44.5 \dnt{11.1} & 22.3 \dnt{63.1} & 76.8 \dnt{0.8} & 11.2 \dnt{55.7} & 81.5 \dnt{2.6} \\
\textit{Clean} & 32.9 \nzt & 48.6 \nzt & 26.3 \nzt & 55.6 \nzt & 85.3 \nzt & 77.6 \nzt & 66.9 \nzt & 84.1 \nzt \\
\midrule
Random-Prepend & 32.5 \dnt{0.3} & 49.8 \upt{1.2} & 29.9 \upt{3.5} & 56.1 \upt{0.5} & 85.2 \dnt{0.2} & 77.8 \upt{0.2} & 78.7 \upt{11.8} & 83.7 \dnt{0.4} \\
Random-Append & \underline{30.9} \dnt{2.0} & \underline{44.7} \dnt{3.9} & 28.3 \upt{2.0} & \textbf{12.3} \dnt{43.3} & \underline{82.5} \dnt{2.9} & 78.3 \upt{0.7} & 73.9 \upt{7.0} & \underline{80.9} \dnt{3.2} \\
Textual UE & 32.1 \dnt{0.7} & 48.5 \dnt{0.1} & 46.6 \upt{20.3} & 53.5 \dnt{2.1} & 85.0 \dnt{0.4} & \textbf{75.4} \dnt{2.2} & 83.1 \upt{16.2} & 84.4 \upt{0.3} \\
MEM-3 & 33.3 \upt{0.4} & 49.3 \upt{0.7} & 36.8 \upt{10.5} & 56.3 \upt{0.7} & 85.1 \dnt{0.2} & 77.0 \dnt{0.6} & \underline{73.2} \upt{6.2} & 83.9 \dnt{0.2} \\
MEM-5 & 33.2 \upt{0.3} & 48.7 \upt{0.1} & 35.5 \upt{9.1} & 55.5 \dnt{0.1} & 86.0 \upt{0.7} & 76.8 \dnt{0.8} & 80.2 \upt{13.2} & 84.0 \dnt{0.1} \\
RegText & 32.3 \dnt{0.6} & 49.3 \upt{0.7} & \underline{28.1} \upt{1.8} & 57.0 \upt{1.4} & 85.6 \upt{0.3} & 77.6 \nzt & 78.1 \upt{11.2} & 84.1 \nzt \\
\specialrule{\lightrulewidth}{\aboverulesep}{0pt}
\rowcolor{OursRow}
\MethodName{} (Ours) & \textbf{27.9} \dnt{5.0} & \textbf{23.1} \dnt{25.5} & \textbf{19.1} \dnt{7.2} & \underline{16.5} \dnt{39.1} & \textbf{70.8} \dnt{14.5} & \underline{76.3} \dnt{1.3} & \textbf{48.0} \dnt{18.9} & \textbf{72.0} \dnt{12.1} \\
\specialrule{\lightrulewidth}{0pt}{\belowrulesep}
\addlinespace[3pt]
& \multicolumn{4}{c}{\textbf{HumanEval}}
& \multicolumn{4}{c}{\textbf{MedQA}} \\
\cmidrule(lr){2-5}\cmidrule(lr){6-9}
\textbf{Method}
& Gemma-12B & \mbox{Llama-3.1-8B} & Phi-4 & Qwen3-4B
& GLM-4-9B & Phi-4 & Qwen3-4B & Qwen3-14B \\
\midrule
\textit{Zero-Shot} & 87.5 \dnt{9.4} & 59.4 \dnt{6.2} & 71.9 \dnt{9.4} & 62.5 \dnt{15.6} & 45.6 \dnt{7.9} & 17.5 \dnt{44.2} & 47.8 \dnt{9.2} & 64.3 \dnt{3.1} \\
\textit{Clean} & 96.9 \nzt & 65.6 \nzt & 81.3 \nzt & 78.1 \nzt & 53.6 \nzt & 61.7 \nzt & 57.0 \nzt & 67.5 \nzt \\
\midrule
Random-Prepend & \underline{81.3} \dnt{15.6} & 56.3 \dnt{9.4} & \underline{78.1} \dnt{3.1} & 37.5 \dnt{40.6} & 53.2 \dnt{0.4} & 53.2 \dnt{8.5} & 53.6 \dnt{3.4} & 68.4 \upt{0.9} \\
Random-Append & 87.5 \dnt{9.4} & \underline{43.8} \dnt{21.9} & \underline{78.1} \dnt{3.1} & 62.5 \dnt{15.6} & \underline{51.0} \dnt{2.6} & \textbf{52.8} \dnt{8.9} & 50.2 \dnt{6.8} & 65.4 \dnt{2.1} \\
Textual UE & 84.4 \dnt{12.5} & 56.3 \dnt{9.4} & \underline{78.1} \dnt{3.1} & 37.5 \dnt{40.6} & 53.3 \dnt{0.2} & 63.2 \upt{1.5} & \underline{46.7} \dnt{10.2} & \underline{64.8} \dnt{2.7} \\
MEM-3 & 87.5 \dnt{9.4} & 59.4 \dnt{6.2} & \textbf{75.0} \dnt{6.3} & 37.5 \dnt{40.6} & 53.2 \dnt{0.4} & \underline{52.9} \dnt{8.8} & 53.8 \dnt{3.1} & 67.3 \dnt{0.2} \\
MEM-5 & 84.4 \dnt{12.5} & 62.5 \dnt{3.1} & \underline{78.1} \dnt{3.1} & 40.6 \dnt{37.5} & 53.5 \dnt{0.1} & 55.9 \dnt{5.7} & 53.6 \dnt{3.4} & 67.2 \dnt{0.2} \\
RegText & 90.6 \dnt{6.3} & 53.1 \dnt{12.5} & \underline{78.1} \dnt{3.2} & \underline{34.4} \dnt{43.7} & 53.5 \dnt{0.1} & 57.2 \dnt{4.5} & 47.3 \dnt{9.7} & 67.0 \dnt{0.5} \\
\specialrule{\lightrulewidth}{\aboverulesep}{0pt}
\rowcolor{OursRow}
\MethodName{} (Ours) & \textbf{78.1} \dnt{18.8} & \textbf{40.6} \dnt{25.0} & \textbf{75.0} \dnt{6.3} & \textbf{31.3} \dnt{46.9} & \textbf{47.3} \dnt{6.3} & 53.0 \dnt{8.6} & \textbf{46.5} \dnt{10.5} & \textbf{40.1} \dnt{27.4} \\
\specialrule{\heavyrulewidth}{0pt}{\belowrulesep}
\end{tabularx}
\end{table*}

\subsubsection{Transfer Across Fine-Tuning Configurations}
\begin{figure}[th]
\centering
\includegraphics[width=0.75\columnwidth]{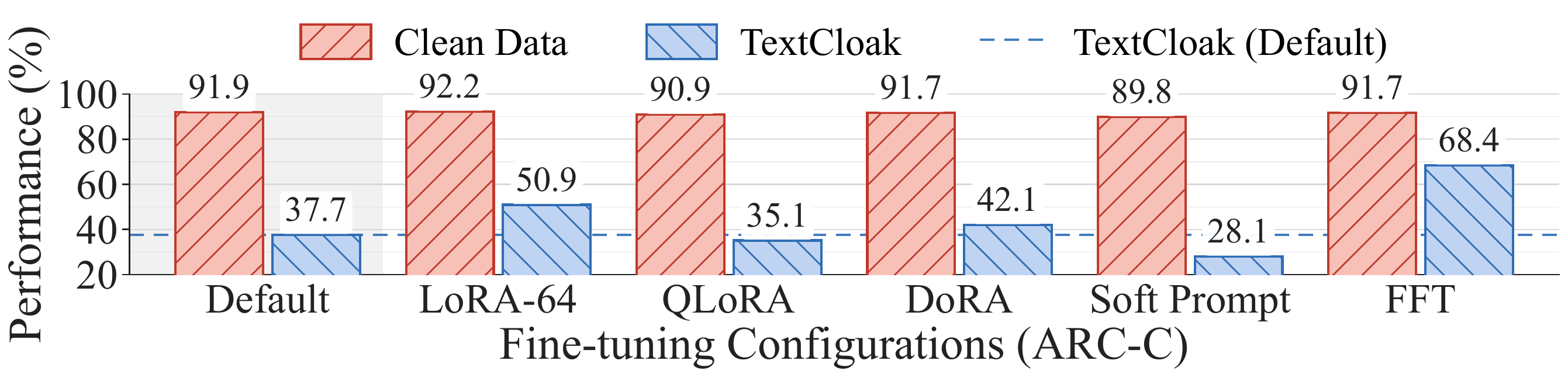}
\caption{Transferability across fine-tuning configurations.}
\label{fig:fine_tuning_transferability}
\end{figure}

We also evaluate the transferability across different fine-tuning configurations, including LoRA-64, QLoRA, DoRA, soft-prompt, full fine-tuning (FFT), as shown in Figure~\ref{fig:fine_tuning_transferability}. Overall, \MethodName{} demonstrates strong transferability across various fine-tuning methods. For instance, \MethodName{} achieves the largest performance degradation on soft prompt while remaining relatively less effective on FFT, which is because fine-tuning with larger parameter updates unlocks more capacity and leads to better adaptation to the protected text.

\Needspace{10\baselineskip}
\subsection{Ablation Study}
\label{sec:ablation}
\begin{wraptable}{r}{0.55\textwidth}
\centering
\small
\caption{Ablation study of \MethodName{}. Variant w/o \OptimizationName{} uses the untrained policy as a paraphraser. The data quality is averaged across three datasets.}
\label{tab:ablation}
\setlength{\tabcolsep}{3pt}
\begin{tabular}{l ccc cc}
\toprule
& \multicolumn{3}{c}{\textbf{Task Performance (\%)}} & \multicolumn{2}{c}{\textbf{Data Quality}} \\
\cmidrule(lr){2-4}\cmidrule(lr){5-6}
\textbf{Variant} & ARC-C & MATH & MedQA & SBERT$\,\uparrow$ & PPL$\,\downarrow$ \\
\specialrule{\lightrulewidth}{\aboverulesep}{0pt}
\rowcolor{OursRow}
\MethodName{} (full) & \textbf{42.7} & \textbf{34.5} & \textbf{41.2} & 0.96 & 12.4 \\
\specialrule{\lightrulewidth}{0pt}{\belowrulesep}
w/o $r_{\mathrm{deg}}$ & 74.4 & 35.3 & 47.0 & 0.96 & 13.5 \\
w/o $r_{\mathrm{sem}}$ & 73.0 & 39.0 & 50.2 & 0.48 & 7.1 \\
w/o $r_{\mathrm{nat}}$ & 61.4 & 35.7 & 47.3 & 0.95 & 11.5 \\
\midrule
w/o $r_{\mathrm{deg}}$, $r_{\mathrm{sem}}$ & 82.4 & 36.0 & 55.9 & 0.64 & 7.7 \\
w/o $r_{\mathrm{deg}}$, $r_{\mathrm{nat}}$ & 66.7 & 36.9 & 52.9 & 0.96 & 12.4 \\
w/o $r_{\mathrm{sem}}$, $r_{\mathrm{nat}}$ & 89.6 & 37.6 & 63.6 & 0.12 & 20.0 \\
\midrule
w/o \OptimizationName{} & 81.8 & 36.3 & 64.7 & 0.91 & 9.6 \\
\bottomrule
\end{tabular}
\end{wraptable}
\emph{RQ3: How do the components of \MethodName{} contribute to its performance?} To investigate this question, we design ablation study to isolate the contribution of each reward as illustrated in Table~\ref{tab:ablation}. The full method achieves the lowest task performance on all three datasets while retaining the highest SBERT score. Removing the degradation reward substantially weakens protection (e.g., 42.7 to 74.4 on ARC-C), and removing \OptimizationName{} reduces the policy to a plain paraphraser with little protective effect, confirming that simulated downstream degradation is the key learning signal. Notably, ablation of the semantic or naturalness reward not only increases the task performance but also deteriorates the data quality. For instance, removing the semantic reward raises task performance on MedQA-USMLE from 41.2 to 50.2 while SBERT drops from 0.96 to 0.48, and removing both the semantic and naturalness rewards further raises it to 63.6 while SBERT collapses to 0.12, which indicates that the utility rewards regularize the policy and potentially mitigate reward hacking. Overall, each component of \MethodName{} is essential for achieving a subtle synthesis of protection and utility.

\subsection{Robustness Analysis}
\label{sec:robustness}
\emph{RQ4: How robust is \MethodName{} against various adaptive attacks, including applying defense strategies and dilution with clean data?} To respond to this research question, we conduct robustness analysis.

\subsubsection{Defense Strategies}
\begin{figure}[th]
\centering
\includegraphics[width=0.75\columnwidth]{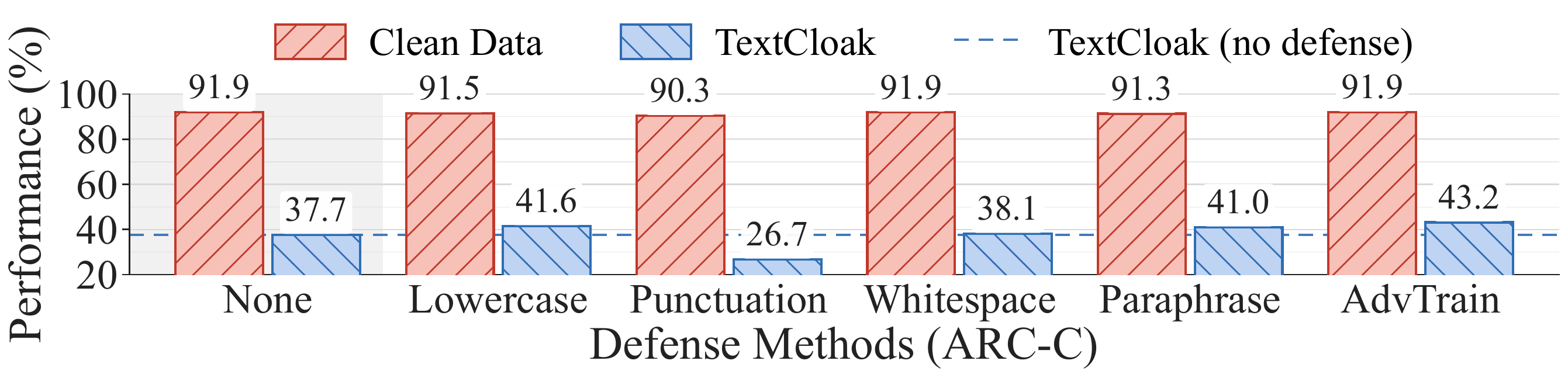}
\caption{Task performance under various defense strategies.}
\label{fig:robustness}
\end{figure}

Unauthorized trainers may attempt to invalidate the protection by transforming the released corpus (e.g., lowercase normalization, punctuation removal, whitespace stripping, and paraphrasing) or adopting adversarial training techniques, as shown in Figure~\ref{fig:robustness}. Generally, the protected text maintains its effectiveness against various defense approaches. Among them, adversarial training is the most effective while punctuation removal is the least effective, suggesting that \MethodName{} provides strong resistance beyond surface-level modifications.

\subsubsection{Data Mixing}
\begin{wrapfigure}{r}{0.5\textwidth}
\centering
\includegraphics[width=0.5\textwidth]{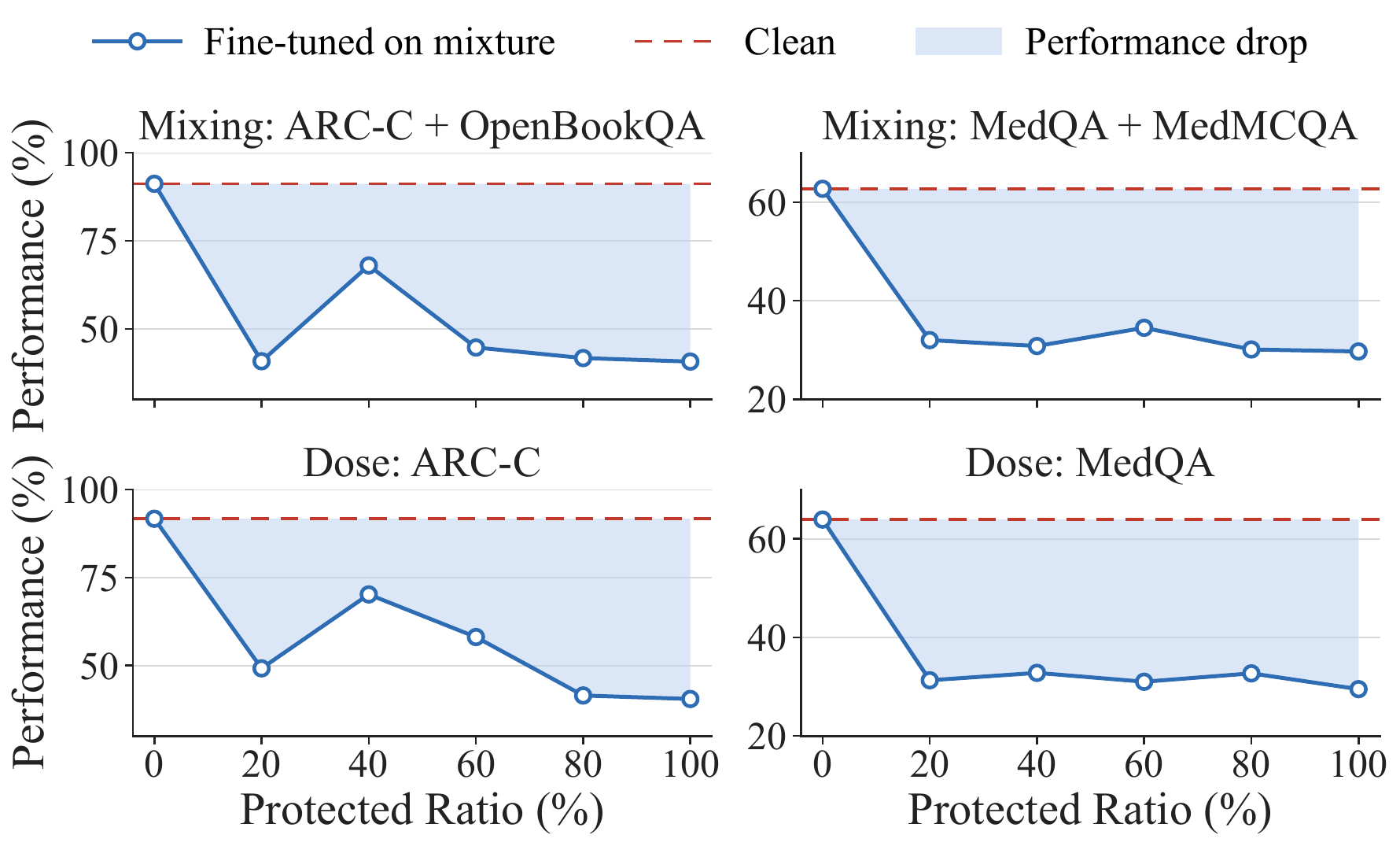}
\caption{Task performance under different proportions of protected text. 1000 samples are randomly selected for the analysis. Top: protected text mixed with clean text from a similar dataset (data mixing). Bottom: protected text mixed with clean copies of the same dataset (dose analysis).}
\label{fig:mixing_dose}
\end{wrapfigure}
Unauthorized trainers may also attempt to mix the protected text with clean text from other sources (e.g., similar datasets or web-crawled corpora) during fine-tuning. To simulate this scenario, we mix protected ARC-C and MedQA data with clean OpenBookQA and MedMCQA data, respectively, and vary the protected ratio, as shown in Figure~\ref{fig:mixing_dose} (top). Remarkably, diluting with clean text from a similar distribution provides little performance recovery. For instance, with only 20\% protected data, the performance drops from 91.2\% to 40.8\% on ARC-C and from 62.7\% to 32.0\% on MedQA-USMLE, nearly identical to the fully protected setting. This suggests that the protective patterns learned by \MethodName{} are highly effective, and clean text from other sources cannot override them during fine-tuning. Consequently, unauthorized trainers cannot recover the utility by mixing in external clean data.

\subsubsection{Dose Analysis} In \emph{extreme cases}, unauthorized trainers may obtain some clean copies of the protected text (e.g., via data leakage or insider attacks) before the defender releases the protected text. We evaluate the robustness of \MethodName{} under different protection doses, as shown in Figure~\ref{fig:mixing_dose} (bottom). Notably, a small dose of protected text is sufficient to induce substantial degradation: with only 20\% protected data, the performance drops from 91.7\% to 49.2\% on ARC-C and from 63.9\% to 31.3\% on MedQA-USMLE, approaching the fully protected setting. Although clean copies of the same instances yield a partial recovery on ARC-C at intermediate doses, the performance converges to the fully protected level once the protected ratio reaches 80\%. Overall, leaked clean copies cannot neutralize the protection unless the unauthorized trainer already possesses nearly the entire clean corpus.

\subsection{Human Evaluation}
\label{sec:human_evaluation}
\begin{table}[th]
\centering
\small
\caption{Human evaluations on a three-point scale. Scores are averaged across human annotators and LLM judges.}
\label{tab:human_evaluation}
\setlength{\tabcolsep}{6pt}
\begin{tabular}{l ccc}
\toprule
\textbf{Method} & Stealthiness$\,\uparrow$ & Fluency$\,\uparrow$ & Human Utility$\,\uparrow$ \\
\midrule
\textit{Clean} & 2.92\,\stdv{0.07} & 2.93\,\stdv{0.05} & 2.99\,\stdv{0.01} \\
\midrule
Textual UE & 2.25\,\stdv{0.12} & 2.03\,\stdv{0.27} & 1.65\,\stdv{1.19} \\
MEM-5 & 1.35\,\stdv{0.56} & 1.57\,\stdv{0.31} & 1.94\,\stdv{0.28} \\
\specialrule{\lightrulewidth}{\aboverulesep}{0pt}
\rowcolor{OursRow}
\MethodName{} (Ours) & \textbf{2.82}\,\stdv{0.11} & \textbf{2.94}\,\stdv{0.04} & \textbf{2.91}\,\stdv{0.08} \\
\specialrule{\heavyrulewidth}{0pt}{\belowrulesep}
\end{tabular}
\end{table}
\emph{RQ5: Do the protected examples preserve language quality and overall utility for legitimate users?} To answer this question, we sample 20 samples from each dataset and then employ three human annotators and three LLM judges to evaluate the stealthiness, fluency, and utility of the protected text. The evaluation protocol is provided in Appendix~\ref{app:human_rubrics}. As showcased in Table~\ref{tab:human_evaluation}, \MethodName{} achieves comparable scores to the clean examples and significantly outperforms the SOTA methods across all metrics. Significantly, the protected text generated by \MethodName{} retains fluency and human utility similar to the original text, indicating that the protected text is practically usable while less compromising the user experience.

\subsection{Case Study}
\label{sec:case_study}
\begin{wrapfigure}{r}{0.55\textwidth}
\centering
\includegraphics[width=0.55\textwidth]{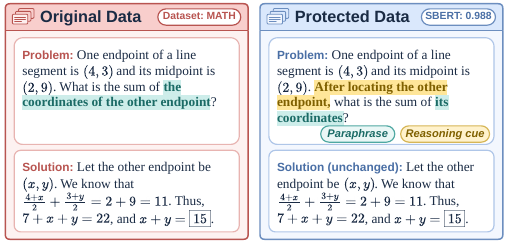}
\caption{Case study.}
\label{fig:case_study}
\end{wrapfigure}
\emph{RQ6: How does \MethodName{} transform clean text into unlearnable examples?} To answer it, we conduct a case study. Figure~\ref{fig:case_study} presents an example of protected text generated by \MethodName{} on MATH. \MethodName{} converts the original text into UE by inserting a reasoning cue (``After locating the other endpoint'') into the problem, which provides a shortcut that undermines the generalization ability of the downstream model. At the same time, the protected text remains semantically faithful (SBERT similarity of 0.988) and linguistically natural, demonstrating the practical effectiveness of \MethodName{} in preventing unauthorized LLM fine-tuning.

\section{Conclusion}
We presented \MethodName{}, an RL-driven framework for protecting textual data against unauthorized LLM fine-tuning. Its generative policy produces semantically faithful and natural unlearnable text, while \OptimizationName{} directly optimizes the policy using downstream degradation measured on fine-tuned surrogate models. Experiments across six datasets and nine LLMs indicate that \MethodName{} consistently impairs unauthorized LLM fine-tuning, transfers across model architectures and training configurations, and remains robust under adaptive attack strategies. These results demonstrate the potential of RL-guided unlearnable text as a practical, model-transferable approach for proactive data protection. Future work will develop sophisticated UEs in high-stakes domains.

\bibliographystyle{abbrvnat}
\nobibliography*
\bibliography{custom}

@String{Computing = "Computing" }

@String{Computer = "{IEEE} Computer" }

@String{Macmillan = "Macmillan" }

@article{brown2020language,
  title={Language models are few-shot learners},
  author={Brown, Tom and Mann, Benjamin and Ryder, Nick and Subbiah, Melanie and Kaplan, Jared D and Dhariwal, Prafulla and Neelakantan, Arvind and Shyam, Pranav and Sastry, Girish and Askell, Amanda and others},
  journal={Advances in neural information processing systems},
  volume={33},
  pages={1877--1901},
  year={2020}
}

@article{shao2024deepseekmath,
  title={Deepseekmath: Pushing the limits of mathematical reasoning in open language models},
  author={Shao, Zhihong and Wang, Peiyi and Zhu, Qihao and Xu, Runxin and Song, Junxiao and Bi, Xiao and Zhang, Haowei and Zhang, Mingchuan and Li, YK and others},
  journal={arXiv preprint arXiv:2402.03300},
  year={2024}
}

@inproceedings{reimers2019sentence,
  title={Sentence-bert: Sentence embeddings using siamese bert-networks},
  author={Reimers, Nils and Gurevych, Iryna},
  booktitle={Proceedings of the 2019 conference on empirical methods in natural language processing and the 9th international joint conference on natural language processing (EMNLP-IJCNLP)},
  pages={3982--3992},
  year={2019}
}

@article{ouyang2022training,
  title={Training language models to follow instructions with human feedback},
  author={Ouyang, Long and Wu, Jeffrey and Jiang, Xu and Almeida, Diogo and Wainwright, Carroll and Mishkin, Pamela and Zhang, Chong and Agarwal, Sandhini and Slama, Katarina and Ray, Alex and others},
  journal={Advances in neural information processing systems},
  volume={35},
  pages={27730--27744},
  year={2022}
}

@inproceedings{li2025generation,
  title={From generation to judgment: Opportunities and challenges of llm-as-a-judge},
  author={Li, Dawei and Jiang, Bohan and Huang, Liangjie and Beigi, Alimohammad and Zhao, Chengshuai and Tan, Zhen and Bhattacharjee, Amrita and Jiang, Yuxuan and Chen, Canyu and Wu, Tianhao and others},
  booktitle={Proceedings of the 2025 Conference on Empirical Methods in Natural Language Processing},
  pages={2757--2791},
  year={2025}
}

@article{zhao2026see,
  title={To See is Not to Learn: Protecting Multimodal Data from Unauthorized Fine-Tuning of Large Vision-Language Model},
  author={Zhao, Chengshuai and Tan, Zhen and Li, Dawei and Yu, Zhiyuan and Liu, Huan},
  journal={arXiv preprint arXiv:2605.14291},
  year={2026}
}

@inproceedings{cao2015towards,
  title={Towards making systems forget with machine unlearning},
  author={Cao, Yinzhi and Yang, Junfeng},
  booktitle={2015 IEEE symposium on security and privacy},
  pages={463--480},
  year={2015},
  organization={IEEE}
}

@inproceedings{carlini2021extracting,
  title={Extracting training data from large language models},
  author={Carlini, Nicholas and Tramer, Florian and Wallace, Eric and Jagielski, Matthew and Herbert-Voss, Ariel and Lee, Katherine and Roberts, Adam and Brown, Tom and Song, Dawn and Erlingsson, Ulfar and others},
  booktitle={30th USENIX security symposium (USENIX Security 21)},
  pages={2633--2650},
  year={2021}
}

@inproceedings{kandpal2022deduplicating,
  title={Deduplicating training data mitigates privacy risks in language models},
  author={Kandpal, Nikhil and Wallace, Eric and Raffel, Colin},
  booktitle={International Conference on Machine Learning},
  pages={10697--10707},
  year={2022},
  organization={PMLR}
}

@inproceedings{huang2021unlearnable,
title={Unlearnable Examples: Making Personal Data Unexploitable},
author={Hanxun Huang and Xingjun Ma and Sarah Monazam Erfani and James Bailey and Yisen Wang},
booktitle={International Conference on Learning Representations},
year={2021},
url={https://openreview.net/forum?id=iAmZUo0DxC0}
}

@inproceedings{li2023make,
  title={Make text unlearnable: Exploiting effective patterns to protect personal data},
  author={Li, Xinzhe and Liu, Ming},
  booktitle={Proceedings of the 3rd Workshop on Trustworthy Natural Language Processing (TrustNLP 2023)},
  pages={249--259},
  year={2023}
}

@inproceedings{wallace2019universal,
  title={Universal adversarial triggers for attacking and analyzing NLP},
  author={Wallace, Eric and Feng, Shi and Kandpal, Nikhil and Gardner, Matt and Singh, Sameer},
  booktitle={Proceedings of the 2019 conference on empirical methods in natural language processing and the 9th international joint conference on natural language processing (EMNLP-IJCNLP)},
  pages={2153--2162},
  year={2019}
}

@inproceedings{wan2023poisoning,
  title={Poisoning language models during instruction tuning},
  author={Wan, Alexander and Wallace, Eric and Shen, Sheng and Klein, Dan},
  booktitle={International Conference on Machine Learning},
  pages={35413--35425},
  year={2023},
  organization={PMLR}
}

@inproceedings{xu2024instructions,
  title={Instructions as backdoors: Backdoor vulnerabilities of instruction tuning for large language models},
  author={Xu, Jiashu and Ma, Mingyu and Wang, Fei and Xiao, Chaowei and Chen, Muhao},
  booktitle={Proceedings of the 2024 Conference of the North American Chapter of the Association for Computational Linguistics: Human Language Technologies (Volume 1: Long Papers)},
  pages={3111--3126},
  year={2024}
}

@inproceedings{yan2024backdooring,
  title={Backdooring instruction-tuned large language models with virtual prompt injection},
  author={Yan, Jun and Yadav, Vikas and Li, Shiyang and Chen, Lichang and Tang, Zheng and Wang, Hai and Srinivasan, Vijay and Ren, Xiang and Jin, Hongxia},
  booktitle={Proceedings of the 2024 Conference of the North American Chapter of the Association for Computational Linguistics: Human Language Technologies (Volume 1: Long Papers)},
  pages={6065--6086},
  year={2024}
}

@article{radford2019language,
  title={Language models are unsupervised multitask learners},
  author={Radford, Alec and Wu, Jeffrey and Child, Rewon and Luan, David and Amodei, Dario and Sutskever, Ilya and others},
  journal={OpenAI blog},
  volume={1},
  number={8},
  pages={9},
  year={2019}
}

@article{clark2018think,
  title={Think you have solved question answering? try arc, the ai2 reasoning challenge},
  author={Clark, Peter and Cowhey, Isaac and Etzioni, Oren and Khot, Tushar and Sabharwal, Ashish and Schoenick, Carissa and Tafjord, Oyvind},
  journal={arXiv preprint arXiv:1803.05457},
  year={2018}
}

@article{hendrycks2021measuring,
  title={Measuring mathematical problem solving with the math dataset},
  author={Hendrycks, Dan and Burns, Collin and Kadavath, Saurav and Arora, Akul and Basart, Steven and Tang, Eric and Song, Dawn and Steinhardt, Jacob},
  journal={arXiv preprint arXiv:2103.03874},
  year={2021}
}

@article{wang2024mmlu,
  title={Mmlu-pro: A more robust and challenging multi-task language understanding benchmark},
  author={Wang, Yubo and Ma, Xueguang and Zhang, Ge and Ni, Yuansheng and Chandra, Abhranil and Guo, Shiguang and Ren, Weiming and Arulraj, Aaran and He, Xuan and Jiang, Ziyan and others},
  journal={Advances in Neural Information Processing Systems},
  volume={37},
  pages={95266--95290},
  year={2024}
}

@inproceedings{lai2017race,
  title={Race: Large-scale reading comprehension dataset from examinations},
  author={Lai, Guokun and Xie, Qizhe and Liu, Hanxiao and Yang, Yiming and Hovy, Eduard},
  booktitle={Proceedings of the 2017 conference on empirical methods in natural language processing},
  pages={785--794},
  year={2017}
}

@article{chen2021evaluating,
  title={Evaluating large language models trained on code},
  author={Chen, Mark and Tworek, Jerry and Jun, Heewoo and Yuan, Qiming and Pinto, Henrique Ponde De Oliveira and Kaplan, Jared and Edwards, Harri and Burda, Yuri and Joseph, Nicholas and Brockman, Greg and others},
  journal={arXiv preprint arXiv:2107.03374},
  year={2021}
}

@article{jin2021disease,
  title={What disease does this patient have? a large-scale open domain question answering dataset from medical exams},
  author={Jin, Di and Pan, Eileen and Oufattole, Nassim and Weng, Wei-Hung and Fang, Hanyi and Szolovits, Peter},
  journal={Applied Sciences},
  volume={11},
  number={14},
  pages={6421},
  year={2021},
  publisher={MDPI}
}

@inproceedings{liu2024multimodal,
  title={Multimodal unlearnable examples: Protecting data against multimodal contrastive learning},
  author={Liu, Xinwei and Jia, Xiaojun and Xun, Yuan and Liang, Siyuan and Cao, Xiaochun},
  booktitle={Proceedings of the 32nd ACM International Conference on Multimedia},
  pages={8024--8033},
  year={2024}
}

@article{yang2025qwen3,
  title={Qwen3 technical report},
  author={Yang, An and Li, Anfeng and Yang, Baosong and Zhang, Beichen and Hui, Binyuan and Zheng, Bo and Yu, Bowen and Gao, Chang and Huang, Chengen and Lv, Chenxu and others},
  journal={arXiv preprint arXiv:2505.09388},
  year={2025}
}

@article{team2025gemma,
  title={Gemma 3 technical report},
  author={Team, Gemma and Kamath, Aishwarya and Ferret, Johan and Pathak, Shreya and Vieillard, Nino and Merhej, Ramona and Perrin, Sarah and Matejovicova, Tatiana and Ram{\'e}, Alexandre and Rivi{\`e}re, Morgane and others},
  journal={arXiv preprint arXiv:2503.19786},
  year={2025}
}

@article{jiang2023mistral,
  title={Mistral 7B},
  author={Jiang, Albert Q and Sablayrolles, Alexandre and Mensch, Arthur and Bamford, Chris and Chaplot, Devendra Singh and Casas, Diego de las and Bressand, Florian and Lengyel, Gianna and Lample, Guillaume and Saulnier, Lucile and others},
  journal={arXiv preprint arXiv:2310.06825},
  year={2023}
}

@article{agarwal2025gpt,
  title={gpt-oss-120b \& gpt-oss-20b model card},
  author={Agarwal, Sandhini and Ahmad, Lama and Ai, Jason and Altman, Sam and Applebaum, Andy and Arbus, Edwin and Arora, Rahul K and Bai, Yu and Baker, Bowen and Bao, Haiming and others},
  journal={arXiv preprint arXiv:2508.10925},
  year={2025}
}

@article{grattafiori2024llama,
  title={The llama 3 herd of models},
  author={Grattafiori, Aaron and Dubey, Abhimanyu and Jauhri, Abhinav and Pandey, Abhinav and Kadian, Abhishek and Al-Dahle, Ahmad and Letman, Aiesha and Mathur, Akhil and Schelten, Alan and Vaughan, Alex and others},
  journal={arXiv preprint arXiv:2407.21783},
  year={2024}
}

@article{abdin2024phi,
  title={Phi-4 technical report},
  author={Abdin, Marah and Aneja, Jyoti and Behl, Harkirat and Bubeck, S{\'e}bastien and Eldan, Ronen and Gunasekar, Suriya and Harrison, Michael and Hewett, Russell J and Javaheripi, Mojan and Kauffmann, Piero and others},
  journal={arXiv preprint arXiv:2412.08905},
  year={2024}
}

@article{glm2024chatglm,
  title={Chatglm: A family of large language models from glm-130b to glm-4 all tools},
  author={Glm, Team and Zeng, Aohan and Xu, Bin and Wang, Bowen and Zhang, Chenhui and Yin, Da and Zhang, Dan and Rojas, Diego and Feng, Guanyu and Zhao, Hanlin and others},
  journal={arXiv preprint arXiv:2406.12793},
  year={2024}
}

@inproceedings{fu2022robust,
title={Robust Unlearnable Examples: Protecting Data Privacy Against Adversarial Learning},
author={Shaopeng Fu and Fengxiang He and Yang Liu and Li Shen and Dacheng Tao},
booktitle={International Conference on Learning Representations},
year={2022},
url={https://openreview.net/forum?id=baUQQPwQiAg}
}

@inproceedings{ren2023transferable,
title={Transferable Unlearnable Examples},
author={Jie Ren and Han Xu and Yuxuan Wan and Xingjun Ma and Lichao Sun and Jiliang Tang},
booktitle={The Eleventh International Conference on Learning Representations },
year={2023},
url={https://openreview.net/forum?id=-htnolWDLvP}
}

@inproceedings{zhang2023unlearnable,
  title={Unlearnable clusters: Towards label-agnostic unlearnable examples},
  author={Zhang, Jiaming and Ma, Xingjun and Yi, Qi and Sang, Jitao and Jiang, Yu-Gang and Wang, Yaowei and Xu, Changsheng},
  booktitle={2023 IEEE/CVF Conference on Computer Vision and Pattern Recognition (CVPR)},
  pages={3984--3993},
  year={2023},
  organization={IEEE}
}

@inproceedings{java2025towards,
  title={Towards operationalizing right to data protection},
  author={Java, Abhinav and Shahid, Simra and Agarwal, Chirag},
  booktitle={Proceedings of the 2025 Conference of the Nations of the Americas Chapter of the Association for Computational Linguistics: Human Language Technologies (Volume 1: Long Papers)},
  pages={8191--8205},
  year={2025}
}

@article{liu2024survey,
  title={A survey of text watermarking in the era of large language models},
  author={Liu, Aiwei and Pan, Leyi and Lu, Yijian and Li, Jingjing and Hu, Xuming and Zhang, Xi and Wen, Lijie and King, Irwin and Xiong, Hui and Yu, Philip},
  journal={ACM Computing Surveys},
  volume={57},
  number={2},
  pages={1--36},
  year={2024},
  publisher={ACM, New York, NY}
}

@inproceedings{zhang2024remark,
  title={$\{$REMARK-LLM$\}$: A robust and efficient watermarking framework for generative large language models},
  author={Zhang, Ruisi and Hussain, Shehzeen Samarah and Neekhara, Paarth and Koushanfar, Farinaz},
  booktitle={33rd USENIX Security Symposium (USENIX Security 24)},
  pages={1813--1830},
  year={2024}
}

@article{yao2024large,
  title={Large language model unlearning},
  author={Yao, Yuanshun and Xu, Xiaojun},
  journal={Advances in Neural Information Processing Systems},
  volume={37},
  pages={105425--105475},
  year={2024}
}

@article{liu2024expshield,
  title={ExpShield: Safeguarding Web Text from Unauthorized Crawling and LLM Exploitation},
  author={Liu, Ruixuan and Tran, Toan and Wang, Tianhao and Hu, Hongsheng and Wang, Shuo and Xiong, Li},
  journal={arXiv preprint arXiv:2412.21123},
  year={2024}
}

@article{li2026versatile,
  title={Versatile transferable unlearnable example generator},
  author={Li, Zhihao and Cai, Jiale and Xu, Gezheng and Zheng, Hao and Li, Qiuyue and Zhou, Fan and Yang, Shichun and Ling, Charles and Wang, Boyu},
  journal={Advances in Neural Information Processing Systems},
  volume={38},
  pages={17495--17522},
  year={2026}
}

@inproceedings{zhao2026chain,
  title={Is chain-of-thought reasoning of llms a mirage? a data distribution lens},
  author={Zhao, Chengshuai and Tan, Zhen and Ma, Pingchuan and Li, Dawei and Jiang, Bohan and Wang, Yancheng and Yang, Yingzhen and Liu, Huan},
  booktitle={Findings of the Association for Computational Linguistics: ACL 2026},
  pages={15231--15261},
  year={2026}
}

@article{steinhardt2017certified,
  title={Certified defenses for data poisoning attacks},
  author={Steinhardt, Jacob and Koh, Pang Wei W and Liang, Percy S},
  journal={Advances in neural information processing systems},
  volume={30},
  year={2017}
}

@inproceedings{fan2022survey,
  title={A survey on data poisoning attacks and defenses},
  author={Fan, Jiaxin and Yan, Qi and Li, Mohan and Qu, Guanqun and Xiao, Yang},
  booktitle={2022 7th IEEE International Conference on Data Science in Cyberspace (DSC)},
  pages={48--55},
  year={2022},
  organization={IEEE}
}

@article{jayaraman2026permissioned,
  title={Permissioned llms: Enforcing access control in large language models},
  author={Jayaraman, Bargav and Marathe, Virendra and Mozaffari, Hamid and Shen, William and Kenthapadi, Krishnaram},
  journal={Advances in Neural Information Processing Systems},
  volume={38},
  pages={81743--81773},
  year={2026}
}

@article{wang2023far,
  title={How far can camels go? exploring the state of instruction tuning on open resources},
  author={Wang, Yizhong and Ivison, Hamish and Dasigi, Pradeep and Hessel, Jack and Khot, Tushar and Chandu, Khyathi and Wadden, David and MacMillan, Kelsey and Smith, Noah A and Beltagy, Iz and others},
  journal={Advances in Neural Information Processing Systems},
  volume={36},
  pages={74764--74786},
  year={2023}
}

@inproceedings{lightman2024let,
  title={Let's verify step by step},
  author={Lightman, Hunter and Kosaraju, Vineet and Burda, Yuri and Edwards, Harrison and Baker, Bowen and Lee, Teddy and Leike, Jan and Schulman, John and Sutskever, Ilya and Cobbe, Karl},
  booktitle={International Conference on Learning Representations},
  volume={2024},
  pages={39578--39601},
  year={2024}
}

\appendix

\section{Experimental Details}
\label{app:detailed_setting}
This section supplements Section~\ref{sec:experimental_setup} with the details needed to reproduce our experiments. Appendix~\ref{app:datasets} describes the datasets and the sampled data splits, and Appendix~\ref{app:implementation_details} lists the implementation details and hyperparameters of \MethodName{}.

\subsection{Datasets}
\label{app:datasets}

This section introduces the six datasets, describes how each example is converted into an instruction-response pair for fine-tuning, and explains how the data are sampled and split.

\subsubsection{Description and Processing}
Every example is cast into an instruction-response pair $(x_i, y_i)$, where the input $x_i$ states the task and the target $y_i$ is the expected answer. For data protection, only $x_i$ is rewritten, and $y_i$ is kept unchanged; for multiple-choice datasets, the answer candidates are also left intact so that the options remain identical across clean and protected text.
\begin{itemize}[leftmargin=*]
    \item \textbf{ARC-Challenge}~\citep{clark2018think} consists of grade-school science questions with four answer options. The question and options form the input, and the correct option is the target response.
    \item \textbf{MATH}~\citep{hendrycks2021measuring} consists of competition-level problems spanning algebra, geometry, probability, and number theory. The problem statement is the input, and the final answer is the target, which is normalized before exact-match comparison.
    \item \textbf{MMLU-Pro}~\citep{wang2024mmlu} consists of multiple-choice questions across diverse disciplines with up to ten options. The question and options form the input, and the correct option is the target response.
    \item \textbf{RACE}~\citep{lai2017race} consists of reading-comprehension questions from English examinations. The passage, question, and options form the input, and the correct option is the target response.
    \item \textbf{HumanEval}~\citep{chen2021evaluating} consists of Python programming problems. The function signature and docstring form the input, and the reference implementation is the target; correctness is checked by executing the provided unit tests.
    \item \textbf{MedQA-USMLE}~\citep{jin2021disease} consists of medical exam questions in the style of the United States Medical Licensing Examination with four options. The question and options form the input, and the correct option is the target response.
\end{itemize}

\subsubsection{Sampling and Splits}
\begin{wraptable}{r}{0.5\textwidth}
\centering
\caption{Statistics of the sampled data splits.}
\label{tab:app_splits}
\small
\setlength{\tabcolsep}{4pt}
\begin{tabular}{l rrr}
\toprule
\textbf{Dataset} & \textbf{Training} & \textbf{Validation} & \textbf{Testing} \\
\midrule
ARC-Challenge & 1,119 & 299 & 1,172 \\
MATH & 1,200 & 750 & 1,500 \\
MMLU-Pro & 1,200 & 1,000 & 1,500 \\
RACE & 1,200 & 1,000 & 4,934 \\
HumanEval & 100 & 32 & 32 \\
MedQA-USMLE & 1,200 & 1,017 & 1,273 \\
\bottomrule
\end{tabular}
\end{wraptable}
We consider six datasets and nine target LLMs, so fine-tuning and evaluating on the full datasets is prohibitively costly. Following common practice in LLM post-training~\citep{wang2023far,lightman2024let}, we randomly sample a subset of each dataset, as summarized in Table~\ref{tab:app_splits}. The three splits play distinct roles. The \emph{training} split is the corpus to be protected: the policy is trained on it, and its protected version is released for unauthorized fine-tuning. The \emph{validation} split serves as the pool for the clean held-out set $\mathcal{D}_{\mathrm{val}}$ in Equation~\ref{eq:degradation_score}; at each reward computation, $\mathcal{D}_{\mathrm{val}}$ is a mini-batch randomly drawn from this split, with 8 examples for most datasets and 4 for HumanEval given its small size. The \emph{testing} split is held out and used only to evaluate the fine-tuned target LLMs.

\subsection{Implementation Details}
\label{app:implementation_details}

\subsubsection{Policy and Surrogate Models}
The generative policy is initialized from \path{meta-llama/Meta-Llama-3-8B} and the surrogate is \path{Qwen/Qwen3-8B}. Both models run in bfloat16, and LoRA adapters with rank 8, scaling factor 32, and dropout 0.05 are applied to all linear layers while the backbone parameters remain frozen.

\subsubsection{Policy Optimization}
The policy adapter is optimized with AdamW using a learning rate of $1\times10^{-5}$, weight decay $1\times10^{-4}$, and maximum gradient norm 1.0. The GRPO clipping radius is $\varepsilon=0.2$, and advantages are standardized with a numerical constant of $10^{-8}$. The rollout temperature decreases from 1.0 in the first epoch to 0.5 in the second epoch with nucleus probability fixed at 1.0.

\subsubsection{Simulated Fine-Tuning}
For each candidate, we restore the same surrogate and optimizer state, apply one LoRA fine-tuning step on the candidate batch, and evaluate the updated surrogate on the clean validation mini-batch $\mathcal{D}_{\mathrm{val}}$ (Appendix~\ref{app:datasets}). The surrogate uses AdamW with learning rate $2\times10^{-5}$, weight decay $10^{-3}$, cosine decay to $2\times10^{-7}$, and maximum gradient norm 1.0.

\subsubsection{Utility Scorers}
Semantic fidelity is computed with the SBERT checkpoint \path{sentence-transformers/all-MiniLM-L6-v2}, and linguistic naturalness with the GPT-2 (\path{openai-community/gpt2}).

\section{Additional Experimental Results}
\label{app:additional_results}
This section reports the variance across random seeds (Appendix~\ref{app:std}) and the training and inference cost of \MethodName{} (Appendix~\ref{app:efficiency}).

\begin{table*}[th]
\centering
\small
\caption{Task performance (\%) in Table~\ref{tab:main_results} reported as mean $\pm$ standard deviation over three random seeds.}
\label{tab:app_std}
\setlength{\tabcolsep}{9pt}
\begin{tabular}{l cccccc}
\toprule
\textbf{Method} & ARC-C & MATH & MMLU-Pro & RACE & HumanEval & MedQA \\
\midrule
\textit{Zero-Shot} & 78.7\,\stdv{0.2} & 20.7\,\stdv{0.4} & 36.3\,\stdv{0.3} & 76.6\,\stdv{0.0} & 74.0\,\stdv{1.8} & 63.6\,\stdv{0.0} \\
Clean & 91.9\,\stdv{0.1} & 38.5\,\stdv{0.7} & 49.1\,\stdv{0.3} & 89.3\,\stdv{1.2} & 80.2\,\stdv{4.8} & 67.3\,\stdv{1.9} \\
\midrule
Random-Prepend & 91.3\,\stdv{0.2} & 38.1\,\stdv{0.9} & 49.6\,\stdv{3.0} & 88.0\,\stdv{0.2} & 75.0\,\stdv{5.4} & 63.9\,\stdv{0.8} \\
Random-Append & 91.2\,\stdv{0.9} & 36.6\,\stdv{1.8} & 47.9\,\stdv{0.4} & 87.6\,\stdv{0.3} & 74.0\,\stdv{4.8} & 62.8\,\stdv{0.2} \\
Textual UE & 91.4\,\stdv{0.3} & 38.3\,\stdv{0.7} & 47.4\,\stdv{2.9} & 87.1\,\stdv{0.1} & 77.1\,\stdv{4.8} & 63.4\,\stdv{0.6} \\
MEM-3 & 90.3\,\stdv{0.4} & 38.9\,\stdv{1.2} & 50.5\,\stdv{2.3} & 87.2\,\stdv{0.3} & 74.0\,\stdv{7.2} & 64.7\,\stdv{0.8} \\
MEM-5 & 90.8\,\stdv{0.4} & 36.5\,\stdv{1.0} & 49.9\,\stdv{3.3} & 88.3\,\stdv{0.2} & 74.0\,\stdv{6.5} & 63.2\,\stdv{0.5} \\
RegText & 91.3\,\stdv{0.1} & 35.3\,\stdv{1.3} & 48.8\,\stdv{2.9} & 86.1\,\stdv{1.2} & 79.2\,\stdv{3.6} & 64.6\,\stdv{0.4} \\
\specialrule{\lightrulewidth}{\aboverulesep}{0pt}
\rowcolor{OursRow}
\MethodName{} (Ours) & 37.7\,\stdv{6.9} & 34.5\,\stdv{1.4} & 39.9\,\stdv{5.9} & 83.8\,\stdv{1.0} & 72.9\,\stdv{4.8} & 41.2\,\stdv{4.6} \\
\specialrule{\heavyrulewidth}{0pt}{\belowrulesep}
\end{tabular}
\end{table*}

\subsection{Variance Across Random Seeds}
\label{app:std}

Table~\ref{tab:app_std} complements Table~\ref{tab:main_results} with standard deviations over three random seeds. The variance of \MethodName{} is small on MATH and RACE, and HumanEval shows the largest variance for all baselines given its small test split of 32 problems. \MethodName{} varies more on ARC-Challenge, MMLU-Pro, and MedQA-USMLE, where the protected models are most strongly degraded; nevertheless, even its mean plus one standard deviation remains \emph{far below} the best baseline, e.g., 44.6 versus 90.3 on ARC-Challenge, so the conclusions in Section~\ref{sec:overall_Performance} are robust to seed variation.

\subsection{Efficiency Analysis}
\label{app:efficiency}

Table~\ref{tab:app_efficiency} reports the cost of \MethodName{} on each dataset on a single NVIDIA A100 GPU, covering both \OptimizationName{} training and the inference that converts the clean training split into its protected version. The number of training samples counts the examples processed over all epochs. Training cost is dominated by the inner-loop surrogate fine-tuning and grows with sequence length, so RACE, whose inputs contain long passages, and MATH are the most expensive, whereas ARC-Challenge and HumanEval are the cheapest; in all cases, training completes within a day. Once trained, the policy protects an entire training split with a single generation pass, which takes under ten minutes on every dataset, i.e., 0.1 to 1.6 seconds per example, amortizing the training cost over the entire corpus.

\begin{table}[th]
\centering
\caption{Training and inference cost of \MethodName{} on a single NVIDIA A100 GPU. Inference converts the clean training split into protected text with the trained policy.}
\label{tab:app_efficiency}
\footnotesize
\setlength{\tabcolsep}{2.6pt}
\begin{tabular}{l rrr rrr}
\toprule
& \multicolumn{3}{c}{\textbf{Training}} & \multicolumn{3}{c}{\textbf{Inference}} \\
\cmidrule(lr){2-4}\cmidrule(lr){5-7}
\textbf{Dataset} & \# Samples & Time & s/sample & \# Samples & Time & s/sample \\
\midrule
ARC-Challenge & 2,238 & 2 h 35 m & 4.2 & 1,119 & 132 s & 0.12 \\
MATH & 2,400 & 15 h 07 m & 22.7 & 1,200 & 182 s & 0.15 \\
MMLU-Pro & 2,400 & 11 h 23 m & 17.1 & 1,200 & 221 s & 0.18 \\
RACE & 2,400 & 19 h 36 m & 29.4 & 1,200 & 560 s & 0.47 \\
HumanEval & 800 & 1 h 26 m & 6.5 & 100 & 156 s & 1.56 \\
MedQA-USMLE & 2,400 & 10 h 12 m & 15.3 & 1,200 & 269 s & 0.22 \\
\bottomrule
\end{tabular}
\end{table}

\section{Human and LLM Evaluation}
\label{app:human_rubrics}

For each dataset, we randomly sample 20 examples and compare the clean text with the protected text produced by Textual UE, MEM-5, and \MethodName{}. Each example is rated by three human annotators and three LLM judges (\path{gemini-3.5-flash}, \path{claude-opus-4-8}, and \path{gpt-5.5}) on stealthiness, text fluency, and human utility, using the three-point rubric in Table~\ref{tab:app_human_rubric}, where higher scores indicate better quality. Raters see the clean input, the protected input, and the unchanged target, and rate only the protected input from the perspective of a legitimate user who must complete the task; the target only lets raters verify that the input remains answerable. The method identity is withheld, the order of the clean and protected versions is randomized, and each dimension is scored independently. Scores are averaged across raters, examples, and datasets. The LLM judge prompt is given in Appendix~\ref{app:llm_judge_prompt}.

\begin{table*}[th]
\centering
\caption{Human evaluation rubric for protected textual examples.}
\label{tab:app_human_rubric}
\footnotesize
\renewcommand{\arraystretch}{1.1}
\begin{tabular}{>{\raggedright\arraybackslash}p{0.16\textwidth}
                >{\raggedright\arraybackslash}p{0.33\textwidth}
                >{\raggedright\arraybackslash}p{0.43\textwidth}}
\toprule
\textbf{Dimension} & \textbf{Definition} & \textbf{Criteria} \\
\midrule
\textbf{Stealthiness} &
The degree to which the text shows no sign of having been manipulated, i.e., it reads as ordinary task text without suspicious insertions, random strings, code-like fragments, irrelevant phrases, or stylistically abnormal content. &
\textbf{1:} The text contains clearly random, gibberish, code-like, irrelevant, or strongly suspicious content that makes the manipulation obvious to a human reader.\par
\textbf{2:} The text remains understandable but includes mild awkwardness, unusual wording, isolated substitutions, or phrase-like fragments that hint at manipulation.\par
\textbf{3:} The text is indistinguishable from naturally written task text, contextually appropriate, and free from noticeable suspicious or irrelevant content. \\
\midrule
\textbf{Text Fluency} &
The degree to which the text is grammatical, coherent, readable, and locally well formed. This dimension focuses on language quality. &
\textbf{1:} The text is grammatically flawed, fragmented, or difficult to read.\par
\textbf{2:} The text is generally readable but contains minor grammatical errors, awkward phrasing, or local coherence issues.\par
\textbf{3:} The text is fluent, coherent, and easy to read. \\
\midrule
\textbf{Human Utility} &
The degree to which a legitimate human user can understand and respond to the intended task from the text without being hindered by ambiguity, missing information, or meaning changes. &
\textbf{1:} The task cannot be answered reliably because essential information is missing, changed, misleading, or ambiguous.\par
\textbf{2:} The task remains answerable, but with noticeable uncertainty caused by mild ambiguity, reduced clarity, or incomplete evidence.\par
\textbf{3:} The task is clearly answerable from the text, with sufficient information for a confident human response.
\\
\bottomrule
\end{tabular}
\end{table*}

\section{Prompt Templates}
\label{app:prompts}
This section provides the two prompt templates used in our experiments: the editing instruction that conditions the UE generative policy and the rating prompt given to the LLM judges.

\subsection{Prompt for the UE Generative Policy}
\label{app:generative_policy_prompt}
The editing instruction $p$ in Equation~\ref{eq:policy_generation} is shown below. It asks the policy to rewrite only the input of every example, forbids changing the target or introducing surface-level artifacts, and permits instance-specific reasoning or procedural framing, so that \OptimizationName{} can discover generalizable patterns instead of a fixed lexical trigger. The same template is used for all datasets and all $K$ group members; datasets differ only in the fields exposed inside \texttt{[BATCH]} (question and options, plus the passage for RACE, the function signature for HumanEval, and the problem alone for MATH).

\begin{tcolorbox}[promptstyle, title=Prompt Template for the UE Generative Policy $\pi_{\phi}$]
\small
\ttfamily
\raggedright
\textbf{[SYSTEM]}\par
You are a data-protection assistant. You rewrite text so that it resists unauthorized model fine-tuning while remaining fully usable and readable for legitimate human readers.\par
\vspace{2pt}
\textbf{[TASK]}\par
You will receive a batch of \{batch\_size\} training examples. Each example has an \texttt{input} field and a fixed \texttt{target} field. Rewrite only the \texttt{input} field of every example to produce a protected version that will be released in place of the original. Your rewrites should introduce subtle, instance-specific reasoning cues, procedural hints, or framing shifts that a model fine-tuned on this batch may latch onto as shortcuts, rather than learning features that generalize to unseen data. Do not solve the task, and do not reveal, hint at, or alter the target answer.\par
\vspace{2pt}
\textbf{[REQUIREMENTS]}\par
1. Preserve the original meaning, task, factual content, and answer options exactly; do not add, remove, or contradict any information needed to solve the task.\par
2. Never modify, leak, or otherwise change the \texttt{target} field; the rewritten input must remain answerable with the same target as the original.\par
3. Keep every rewritten input fluent, grammatical, and natural, as if written by a careful human author.\par
4. Do not introduce random strings, repeated tokens, code-like fragments, non-sequitur insertions, or any artifact that a human reader would find suspicious or out of place.\par
5. You may vary sentence order, phrasing, or add brief instance-specific reasoning or procedural framing, provided the correct answer and task semantics are unchanged.\par
6. Rewrite every example in the batch; return exactly \{batch\_size\} examples, in the same order, matched by \texttt{id}.\par
7. Output only valid JSON, with no text before or after it: a list of objects with fields \texttt{id} and \texttt{protected\_input}.\par
\vspace{2pt}
\textbf{[OUTPUT FORMAT]}\par
[\{"id": \textit{<id>}, "protected\_input": \textit{"<rewritten input>"}\}, ...]\par
\vspace{2pt}
\textbf{[BATCH]}\par
\{batch\_examples\}\par
\vspace{2pt}
\textbf{[OUTPUT]}
\end{tcolorbox}

\subsection{Prompt for LLM Judges}
\label{app:llm_judge_prompt}
LLM judges receive the prompt below, which embeds the rubric text of Table~\ref{tab:app_human_rubric} verbatim so that human and LLM raters are held to an identical standard. Scores are parsed from the JSON output.

\begin{tcolorbox}[promptstyle, title=Prompt Template for LLM Judges]
\small
\ttfamily
\raggedright
\textbf{[SYSTEM]}\par
You are an expert annotator evaluating text released for legitimate downstream use. You will compare a clean example with a protected (rewritten) version of the same example and judge only the protected version's quality from the perspective of a legitimate human reader who must complete the underlying task.\par
\vspace{2pt}
\textbf{[TASK]}\par
Rate the protected input on the three dimensions below, each on a 3-point scale. The target response is provided only for reference; it is unchanged and is not itself being rated.\par
\vspace{2pt}
\textbf{[DIMENSIONS AND CRITERIA]}\par
\{rubric\} \textit{(the definitions and 1--3 criteria for stealthiness, text fluency, and human utility in Table~\ref{tab:app_human_rubric})}\par
\vspace{2pt}
\textbf{[INPUT]}\par
Task domain: \{task\_domain\}\par
Clean input: \{clean\_input\}\par
Protected input: \{protected\_input\}\par
Target response (reference only, not rated): \{target\_response\}\par
\vspace{2pt}
\textbf{[OUTPUT FORMAT]}\par
Output only valid JSON, with no text before or after it:\par
\{"stealthiness": \textit{<1-3>}, "fluency": \textit{<1-3>}, "human\_utility": \textit{<1-3>}, "justification": \textit{"<one or two sentence rationale>"}\}\par
\vspace{2pt}
\textbf{[OUTPUT]}
\end{tcolorbox}

\section{Ethics and Privacy Statement}
\label{app:ethics}
\MethodName{} aims to give data owners control over whether their published text can be exploited by unauthorized LLM fine-tuning. It only modifies data that the owner controls before release, keeps the target responses unchanged, and does not implant backdoors or harmful behaviors; models trained on the protected text merely generalize worse on the protected task. Nevertheless, the technique is dual-use: applied to shared corpora that others are entitled to use, it could degrade legitimate models, so it should be applied only to data whose use the owner has the right to restrict. Protection is not guaranteed against every adaptive adversary (Section~\ref{sec:robustness}) and should complement, rather than replace, access control for sensitive data. All experiments use publicly available benchmarks. Human annotators rated only public benchmark text and were not exposed to personal or harmful content.

\section{Use of Generative AI}
\label{app:generative_ai}

We used GPT-5.6 Sol and Claude Fable 5.1 solely as language polishing tools for proofreading, grammatical correction, and stylistic refinement, analogous to a grammar checker. They did not contribute to the generation of scientific content or ideas.

\end{document}